\documentclass[10pt,journal,compsoc]{IEEEtran}
\ifCLASSOPTIONcompsoc
  \usepackage[nocompress]{cite}
\else
  \usepackage{cite,amssymb,hyperref}
\fi
\ifCLASSINFOpdf
\else
\fi

\usepackage{amsmath}

\newcommand{\mytilde}{\raise.22ex\hbox{$\scriptstyle\mathtt{\sim}$}}

\usepackage[caption=false,font=footnotesize]{subfig}

\usepackage{multirow}
\usepackage{amssymb,tabularx, graphicx, booktabs}

\usepackage[figuresright]{rotating}

\usepackage[boxed]{algorithm2e}

\begin{document}
%
\title{Gamifying Video Object Segmentation}
%
%
%
%

\author{Simone~Palazzo,
        Concetto~Spampinato,~\IEEEmembership{Member,~IEEE} and Daniela~Giordano,~\IEEEmembership{Member,~IEEE}
\IEEEcompsocitemizethanks{\IEEEcompsocthanksitem S. Palazzo, C. Spampinato and D. Giordano are with the Department
of Electrical, Electronics and Computer Engineering, University of Catania, Italy.\protect\\
E-mail: see http://perceive.dieei.unict.it}
\thanks{Manuscript submitted on January 05, 2016}}

\maketitle

\begin{abstract}
Video object segmentation can be considered as one of the most challenging computer vision problems. Indeed, so far, no existing solution is able to effectively deal with the peculiarities of real-world videos, especially in cases of articulated motion and object occlusions; limitations that appear more evident when we compare their performance with the human one. However, manually segmenting objects in videos is largely impractical as it requires a lot of human time and concentration. 
To address this problem, in this paper we propose an interactive video object segmentation method, which exploits, on one hand, the capability of humans to identify correctly objects in visual scenes, and on the other hand, the collective human brainpower to solve challenging tasks. In particular, our method relies on a web game to collect human inputs on object locations, followed by an accurate segmentation phase achieved by optimizing an energy function encoding spatial and temporal constraints between object regions as well as human-provided input. 
Performance analysis carried out on challenging video datasets with some users playing the game demonstrated that our method shows a better trade-off between annotation times and segmentation accuracy than interactive video annotation and automated video object segmentation approaches.

\end{abstract}

\begin{IEEEkeywords}
Interactive video annotation, Games with a purpose, Human in the Loop, Spatio-temporal superpixel segmentation
\end{IEEEkeywords}


\section{Introduction}
\label{sec:introduction}
\IEEEPARstart{T}{he} generation and collection of massive amount of videos has become an easy task due to the progress in low-cost digital imaging systems as well as storage services. Indeed, every day several petabytes of videos are routinely generated for disparate applications ranging from video-surveillance to news broadcasting to entertainment. This is also highlighted in the recent ``Forecast and Methodology 2014-2019'' report\footnote{http://www.cisco.com/c/en/us/solutions/collateral/service-provider/ip-ngn-ip-next-generation-network/white\_paper\_c11-481360.html}  by CISCO that has estimated that consumer internet video traffic will be 80 percent of all consumer Internet traffic in 2019. Nevertheless, this video data deluge needs automated methods able to extract meaningful information for data indexing, preservation and understanding, since it is unrealistic and unfeasible (as stated in the same CISCO report, {\it it would take an individual over 5 million years to watch the amount of video that will cross global IP networks each month in 2019}) to assume people can do this work manually. This is the main reason why all the  existing video datasets have only a few frames annotated.
One of the upstream modules for video understanding is object segmentation, which aims at discriminating accurately foreground objects from the background. 
There is a large (past and present) bulk of literature of methods for video object segmentation. Background modeling/subtraction~\cite{Giordano2015superpixel, HanKDESVM},  motion analysis~\cite{6682905,baieccv2010}, object ranking~\cite{Zhang2013video,Lee:2011:KVO:2355573.2356444} and clustering point tracks~\cite{Ochs2011,broxeccv10} and, recently, combination of CNN-based moving object detectors~\cite{Fragkiadaki_2015_CVPR} are among the most common methods. However, the accuracy and performance of these techniques are still not satisfactory, especially in cases of articulated motion, cluttered scenes and object occlusions. 
Therefore, so far, there exist no valid alternatives to the classic automated video segmentation methods and the only possible solution might be to include effectively and efficiently humans in the analysis and learning process. ``Human in the loop'' is a recent trend in machine learning which tries to learn discriminative patterns by proactively involving people in the annotation process. This is the case of ``games with a purpose'' that channel collective human brainpower through computer games~\cite{vonAhn2008}. The underlying idea is to engage people in solving unconsciously complex tasks while playing computer games. The combination of these games to crowdsourcing strategies may be an extremely powerful tool to solve tasks at large scale. 
While there exist several games with purpose to support automated image analysis~\cite{Morrison:2009,Hacker:2009,vonAhn2004,vonAhn:2006,vonAhn2006}  and some for video tagging~\cite{eps271071,ontogame} (which, however, share the same philosophy of the image annotation ones), to the best of our knowledge, none have been adopted for video object segmentation, which would particularly benefit from this approach as annotating videos, in terms of object segmentation, requires much more human time and concentration than identifying image classes.
Under this scenario, in this paper we propose a human-guided video object segmentation method, built upon a web game and able to effectively and accurately extract moving objects from videostreams by greatly reducing human intervention. 
The contribution of the paper is threefold:
\begin{itemize}
\item First, we present and release a web game to collect human input (in the form of clicks) that can be used with any kind of videos, thus representing a powerful tool for computer vision scientists (and not only) to get their own videos automatically annotated;
\item We propose an interactive video object segmentation approach based on the optimization of an energy function which is able to encode spatio-temporal constraints between object regions as well as human-provided priors. The method can be combined with other sources of human input (e.g., eye-gaze data~\cite{Palazzo2015using}) to annotate automatically videos.
\item We demonstrate that the collective action of players, despite providing noisy and inaccurate data, results in better segmentation accuracy -- with much less human effort --  than state-of-the-art automated video object solutions as well as interactive video annotation methods.
\end{itemize}


\section{Related Work}
\label{sec:related}
Our work shares the same end goal of automated video object segmentation~\cite{Giordano2015superpixel,Barnich2011vibe,Spampinato2014texton,HanKDESVM,Shengcai2010, Zhang2011,Papazoglou2013fast,eccv}, but our approach is more inline with research that places humans in the loop (including games with purpose)~\cite{Deng2013,Maji2012,Sorokin,Parikh2012,Parkash,Parikh,Vijayanarasimhan2014,Salvador2013,Morrison:2009,Hacker:2009,vonAhn2004,vonAhn:2006,vonAhn2006,eps271071,ontogame} and interactive video segmentation~\cite{Badrinarayanan2010,Ochs2011,Fathi2011,Budvytis2011,NSB15,Price2009,Badrinarayanan2013,Li2005}.

Unsupervised video segmentation has gained a lot of attention in the last decades~\cite{Barnich2011vibe,Spampinato2014texton,HanKDESVM, Shengcai2010, Zhang2011} and recently it has been thought mainly in terms of spatio-temporal superpixel modeling~\cite{Giordano2015superpixel,Papazoglou2013fast, eccv}. The key idea behind these methods is the one of grouping pixels which are appearance- and motion-wise--consistent. Despite the performance increase due to superpixel segmentation, all these methods suffer from oversegmentation, especially in cases of camera motion and object occlusions.
Therefore, manual or semi-manual video annotation can be considered the only reliable way for obtaining precise object segmentation, but as argued earlier, this tedious and labor-intensive process is extremely costly and unfeasible at large scale. Semi-supervised video segmentation approaches~\cite{Badrinarayanan2010,Ochs2011,Fathi2011,Budvytis2011,NSB15,Price2009}, that differently from the semi-supervised image segmentation ones (e.g., the popular Grabcut~\cite{grabcut}) have not received much attention, usually require a short intervention by humans in terms of object annotations that are then propagated automatically over time. Most of these methods rely either on optical flow~\cite{Fathi2011} or on temporal connections ~\cite{Price2009, Budvytis2011} between frames usually modeled by Markov chains, but they work well only in simple cases failing mainly in precisely representing object boundaries. In~\cite{Li2005}, the authors propose an interactive video object segmentation method using a 3D graph- cut--based segmentation followed by a tracking-based local refinement. In~\cite{NSB15} video object segmentation is formulated as a spatio-temporal MRF optimization problem, with a cost function including user input, motion and appearance cues, spatio-temporal consistency similarly to the one proposed in this work.  In addition, superpixel segmentation is also largely employed as it allows us to reduce processing time ensuring at the same time spatio-temporal coherency among pixels. In some work~\cite{jain2014}, temporal linking between superpixels is done manually. Despite these methods are able to alleviate human effort for video annotation, at large scale they are ineffective and still time-consuming for human operators.
Another option to support low-level computer vision tasks is to understand how human perform them and to seek how human inference/reasoning can be integrated into computer programs. Examples are the ones that ask people to provide explicitly annotation rationales \cite{Donahue2011} or to elicit the visual features employed to discriminate between image/object classes~\cite{Vedaldi2014,Deng2013}. Nevertheless, unlike computers, humans need incentives, either monetary or for entertainment, to carry out specific tasks. Under this scenario, on-line games represent an effective mechanism to involve people in solving challenging problems. 
Two of the most common approaches exploiting web-games for collecting human feedback for machine learning methods are the ESP Game~\cite{vonAhn2004} and Peekaboom~\cite{vonAhn2006}. Both approaches use the collective intelligence of human brains for gathering key information for image classification. Since their release,  many thousands of people have played them, generating millions of labels. However, these games are devised only for image analysis and, moreover, cannot be played by everyone: for instance, children, who usually are very passionate with games, would have difficulty in getting engaged by them.  In addition, to the best of our knowledge, there exist no games employed for supporting video object segmentation, except the one used in this work.

The approach proposed in this paper draws inspiration from both interactive video object segmentation approaches and human-computation using web-games combining both strategies in a smart way for accurate object segmentation in videos at large scale. More specifically, we propose an interactive video segmentation approach formulated as a spatio-temporal superpixel labeling by taking into account user input and spatio-temporal consistency of motion and appearance features. In addition, user feedback is gathered through a web game in the form of clicks, instead of strokes (as in most of the interactive video object segmentation methods),  leveraging on multiple users to obtain more consistent and structured feedback for automated segmentation. Also, the web-game is designed to being playable by any person of any age, thus increasing its possible audience (and with it the amount of gathered data) and improving the accuracy of the generated object segmentations.

\section{Method}
\label{sec:methods}
The proposed interactive video object segmentation approach can be seen as a two-step spatio-temporal MRF optimization problem: the first one with a cost function exploiting spatial information at the frame level and encoding user input and appearance cues in order to extract homogeneous object regions in video frames; and the second one enforcing spatio-temporal consistency between the segmented object regions in consecutive frames, thus refining the preliminary segmentation. 
Three are the main modules of the whole approach:
\begin{itemize}
 \item \emph{The game}: The starting point of the whole process, our game is thought to gather user clicks in correspondence of objects of interest in videos. The game is designed to be challenging and competitive, so that users are encouraged to play: while this helps keeping the competition between users, game difficulty often reflects on the noisiness of the generated data.
 \item \emph{Superclick extraction}: The initial stage of our algorithm converts the noisy set of clicks into a set of more accurate ``clicked superpixels'', or \emph{superclicks}. Posing the problem in terms of superpixels rather than pixels 1) reduces the  numerical complexity of the task and, 2) enforces spatial coherency between clicked object regions. On top of this viewpoint, we identify and group together superclicks through MRF optimization. 
 \item \emph{Temporal smoothing}: Single-frame superclick extraction produces a fairly accurate segmentation of the objects in the scene, however it ignores temporal consistency between frames which can be exploited to further improve the segmentation. Based on the superclicks extracted from a span of consecutive frames, a three-dimensional (across time) MRF is designed in order to transfer information on the labels assigned to corresponding superpixels at different frames.
\end{itemize}

\subsection{The Game}
\label{sec:game}
We reused the game presented in \cite{6595949} (where users had to perform a similar task but for a different objective, i.e., ground-truth generation) by adapting and modifying it according to our objectives: players are instructed to click on moving objects in a set of videos (one for each game level). Each correct click awards points, and each level is successfully completed if the user sums up a certain amount of points (increasing by level). 

\begin{figure}
 \centering
 \includegraphics[width=0.48\textwidth]{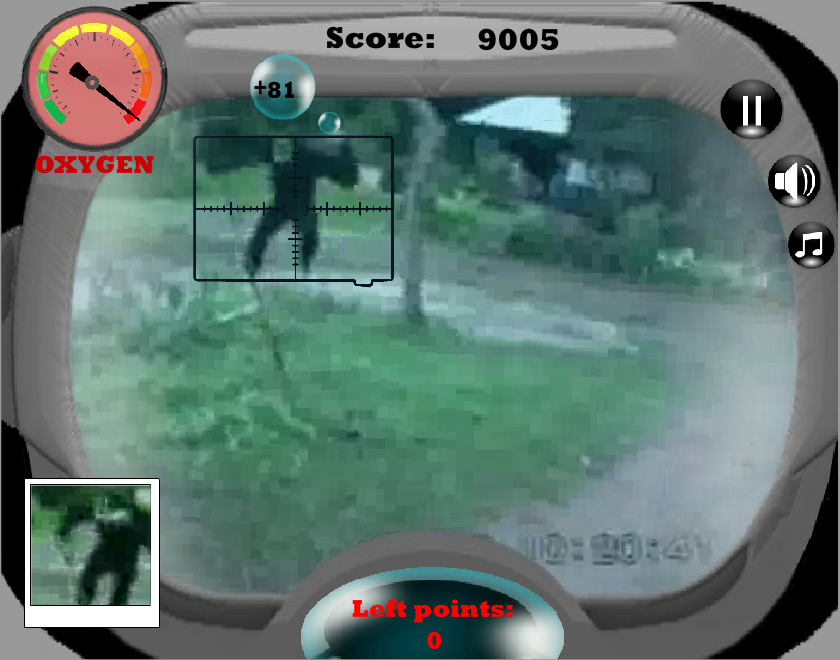}
 \caption{In-game screenshot of the user interface.}
 \label{fig:game_interface}
\end{figure}

\textbf{User interface.} Fig.~\ref{fig:game_interface} shows an example of a typical in-game screenshot. The video for the current level is, of course, the most important element of the interface and takes up most of the space; the current score obtained by the user is shown at the top; the remaining time before the level's end is shown on the top-left corner (indicated by the \textsc{oxygen} icon---a legacy from the original underwater-oriented application of the game), and the number of points needed to pass the current level is at the bottom of the screen. The mouse cursor is shaped like a camera reticle, and at each click the taken ``photo'' is shown  at the bottom-left corner of the screen (this is done for future object classification purposes, which are beyond the scope of this paper). When the user clicks correctly on a target, points are awarded, shown as upward-floating bubbles (``+81'' in the example image). Finally, further option buttons are shown at the top-right corner of the screen.

\textbf{Levels.} Each level in the game is associated to an input video, which is supposed to be at least 30 seconds long at 10 frames per second (if longer, only the first 30 seconds will be shown). In order to pass a level, a certain amount of points must be scored, starting from 4000 at the first level, and increasing by 2000 at each successive level. As the game is actually relatively simple, this increase represents the main challenge, since it makes it more and more difficult to achieve the required points.

Level-video association is done randomly, i.e., in different game sessions, the video ordering is never the same in order to avoid players to know in advance where objects might be located. This is necessary since in games, players often tend to maximize their scores also using tricks.

\begin{figure}
 \centering
 \includegraphics[width=0.4\textwidth]{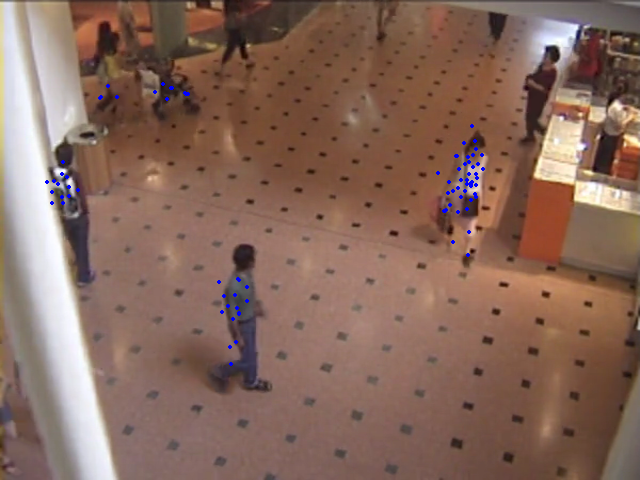} \\ \vspace{0.2cm}
 \includegraphics[width=0.4\textwidth]{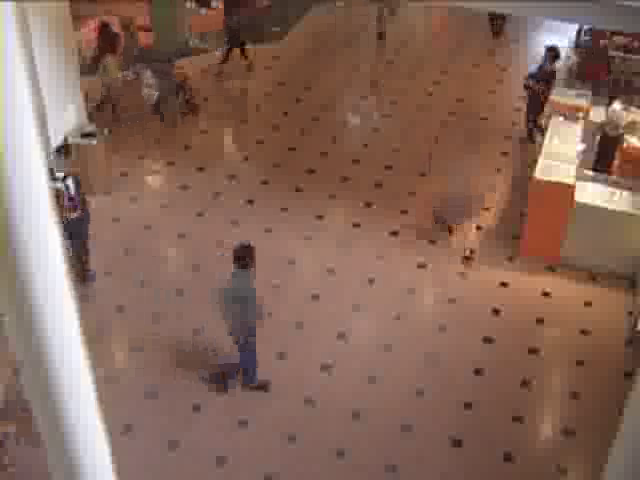}
 \caption{{\bf Left}: user clicks (blue dots) in a frame; {\bf Right}: saliency inhibition by blurring clicked regions.}
 \label{fig:ior_blur}
\end{figure}

One related issue was the \emph{saliency bias} of some objects with respect to others, which caused users to click always on the same objects (the most salient ones) in a scene even if several others were present. To reduce this phenomenon, we applied an \emph{inhibition of return} mechanism by blurring videos in areas where clicks (by all users) accumulate: this reduced the saliency of underlying objects and led users to click on other objects in the scene. Video blurring was performed using all gathered clicks, and not just the current user's, although saliency is partly a subjective process: we found this also helped to avoid gathering too many clicks on objects for which we already had enough data (see paragraph~\ref{sec:results_users_time}). Fig.~\ref{fig:ior_blur} shows an example of the click distribution in a frame and the corresponding blurred version.

\textbf{Points.} As in any gamification process, it is necessary to pose the task as a competitive one, providing the users with a feedback on how good they are with respect to their previous results or their friends. We employ a point-based system to reflect users' performance on the game, and keep an all-time ranking of the best scores.
Points are awarded by clicking correctly on an object of interest, depending on the size of the object and on previous clicks: bigger objects are awarded more points, but successive clicks in the same area earn the user less and less points, according to the formula:
\begin{equation}
P^+ = \frac{A}{20}\left(1 - \frac{t}{10}\right)
\end{equation}
where $P^+$ is number of points earned for a correct click, $A$ is the area in pixels of the clicked object, and $t$ is the number of consecutive clicks within a 30$\times$30 pixels region. In practice, $\frac{A}{20}$ is the maximum score awarded for a click on a certain object, but this score is progressively reduced if the user keeps clicking on the same spot: when salient objects are in the scene, this reduction forces users to vary their clicking pattern to get more points, while at the same time helps to provide data on as many objects as possible. Conversely, users are subtracted points if they click too far from the objects in the visualized frame; the penalty is computed as:
\begin{equation}
P^- = 20t
\end{equation}
where $P^-$ is the amount of points subtracted to the current score due to a wrong click, and $t$ is the number of consecutive clicks falling further than 200 pixels from the closest correct object. Penalties prevent users from clicking randomly across the frame and force them to be as more accurate as possible.

However, to award points to players we need object segmentation (not necessarily highly accurate) on the input videos to tell whether clicks hit or miss objects. In order to have a reference signal --- {\it score video segmentation}--- according to which we assign points to players, we use the \emph{output of the system itself}. When the system is first set up and no data is available yet, the initial video object segmentation is obtained by running a classic background modeling method (\cite{Barnich2011vibe} in our case); although in the beginning this may not be enough to cover all and only objects in the scene, it still provides an adequate base for setting the game up. After users have started to play, the object segmentation is simply updated based on users' clicks by running the algorithm presented in this paper. It is not strictly necessary for {\it score video segmentation} to be extremely accurate: scores are only provided for the benefit of users, in order to keep them interested by means of competition.

\textbf{Click quality.} We also estimate the ``quality'' (in the sense of ``accuracy of clicks with respect to objects'') of the data provided by users while playing the game. Quality scores are computed on a per-level and per-user basis as the fraction of user clicks hitting the objects in the level. We assume that all clicked pixels in a game level by a user gets the same quality score computed as above. 
We could have computed a global quality score for a single game (i.e., the sequence of levels a user plays before completing the game) or for the user, however different levels may return completely different quality scores even within the same game session, due to each video's scene and object characteristics, so a global score would become too generic to describe individual click quality in a level's context.


\subsection{Superclick Extraction}
\label{sec:superclick}

The clicks collected through the web game are used to extract information on the location of objects in the scene for each video frame and to carry out a preliminary object segmentation. We pose the problem as a binary segmentation task (background and foreground) by means of the minimization of an energy function defining the cost of a segmentation. Like some of the most recent methods for video object segmentation~\cite{Giordano2015superpixel,Papazoglou2013fast, eccv}, we use \emph{superpixels} (computed by SLIC \cite{Achanta2012slic}) as basic image parts instead of pixels as they provide two main advantages: 1) reducing the number of variables greatly speeds up the minimization algorithm (the number of variables is scaled down by a factor of 30-50, depending on superpixel settings); 2) the initial segmentation provided by superpixels is usually effective in detecting edges, which allows to simply focus on finding the optimal aggregation, taking boundary detection for granted.

The first processing step, given our target frame $F$, consists of \emph{superclick extraction}, where a \emph{superclick} is the intuitive extension of the concept of clicks to superpixels. This step is necessary to be able to pose the problem in terms of superpixels only, by ``converting'' point data (e.g., clicks) to superpixel-oriented ones. Of course, the principle behind this operation is that superpixels containing clicked pixels should be more likely to be marked as superclicks, which are then converted into constraints for MRF optimization. However, clicks are generally noisy, thus other factors, such as click density, click quality (as defined previously), closeness to other clicked superpixels need to be taken into account for superclick identification.

\begin{figure}
\centering
\includegraphics[width=.24\textwidth]{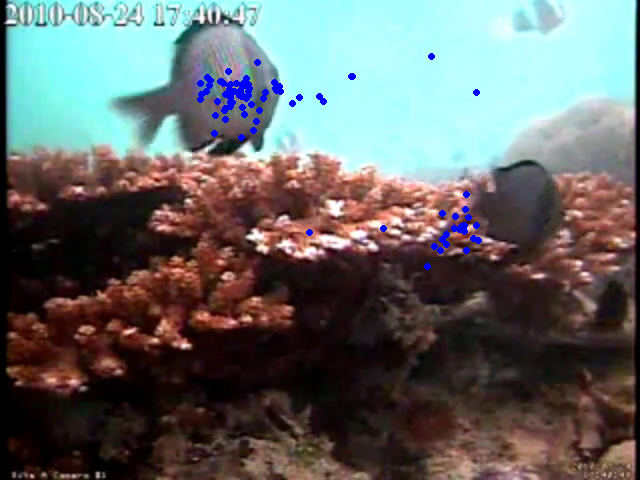}
\includegraphics[width=.24\textwidth]{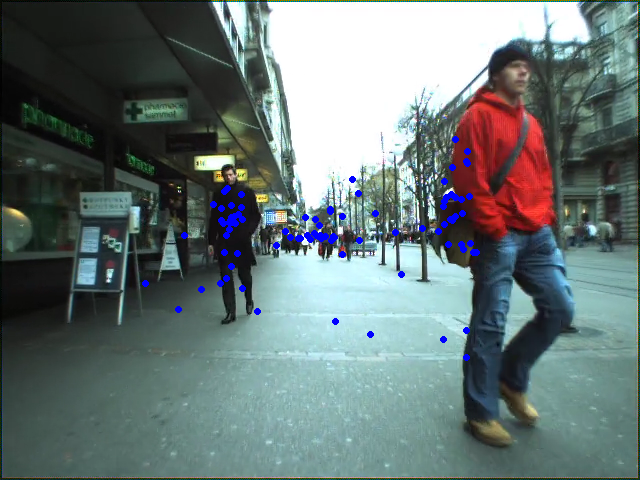}\\ \vspace{0.2cm}
\includegraphics[width=.24\textwidth]{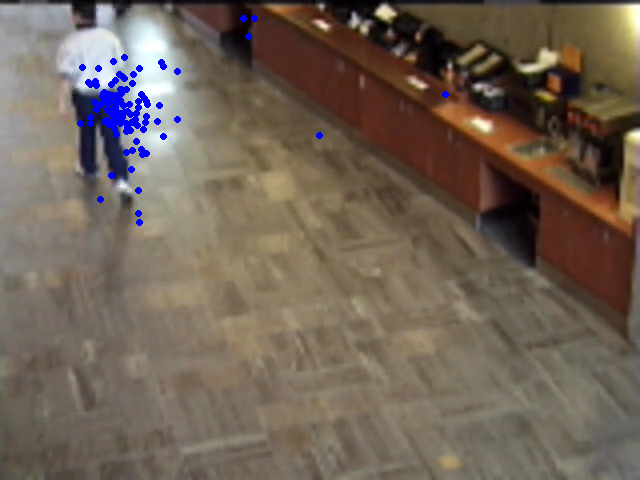}
\includegraphics[width=.24\textwidth]{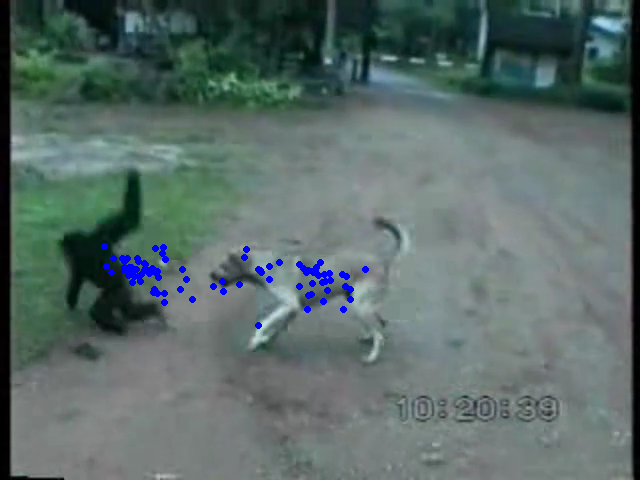}
\caption{Due to users' reaction times, clicks may be delayed with respect to the ``intended'' frame. It is possible to notice that this phenomenon may be more or less evident even within the same image, depending not only on the user but also on the objects in the scene.}
\label{fig:click_delay}
\end{figure}

Before explaining how superclicks are computed, a more basic question is: \emph{what clicks should we use to analyze a certain frame?} Depending on video frame rate and target speed, users' reaction times may introduce a delay which results in a shift between the frame at which the user clicks on an object and the frame at which the user \emph{intended} to click. Fig.~\ref{fig:click_delay} shows a few examples: it is possible to notice that the delay effect is more visible on some videos than others according to mainly objects' speed. Since this issue involves complex biological phenomena~\cite{reaction_times}, which are out of the scope of the paper, we adopt a simple but effective empirical approach: we assume that all clicks are delayed by a constant number of frames for all videos. 
In detail, the results in Sect.~\ref{sec:results} shows that shifting all clicks back by 2 frames (although the optimal delay may vary from 1 to 4 frames which depends on several factors, one above all the video frame rate) represents a good trade-off between accuracy and complexity.

Let $C = \left\{ c_1, c_2, \dots, c_{n_C}\right\} = \left\{ (x_1, y_1), (x_2, y_2), \dots, (x_{n_C}, y_{n_C})\right\}$ be the players' clicks for frame $F$, with corresponding quality scores $Q = \left\{q_{c_1}, q_{c_2}, \dots, q_{c_{N_C}}\right\}$ (each click gets the quality score assigned to the user who did it on a per-level basis). We define a graph-representable energy function~\cite{Kolmogorov2004what} over the set of $F$'s superpixels $S = \left\{s_1, s_2, \dots, s_{n_S} \right\}$, with a cost function able to model the ``clickedness'' of each superpixel independently, and at the same time, to enforce constraints on visual smoothness and click continuity. 
Our main assumptions are:
\begin{enumerate}
 \item Superpixels containing a large number of clicks should be marked as superclicks (and vice versa), i.e., they can be seen as hard constraints for segmentation. 
 \item Clicked pixels should be weighted by the relative quality when evaluating their contribution to a superclick.
 \item Unclicked superpixels which are close to clicked and visually-similar superclicks should be marked as superclicks as well, since they are likely to belong to the same object.
 \item Isolated clicked superpixels (even if in small groups) should be ignored as being likely noise.
\end{enumerate}
Translating these assumptions into energy potentials, we obtain the following cost function for energy optimization:
\begin{equation}
 E_1(\mathcal{L}) = \alpha_1 \sum_{s \in S} V_1(s, l_s, C) + \sum_{(s_1,s_2) \in \mathcal{N}(S)} V_2(s_1,s_2, l_{s_1}, l_{s_2})
 \label{eq:superclick_energy}
\end{equation}
where $\mathcal{L} = \left\{ l_{s_1}, l_{s_2}, \dots, l_{s_{n_S}}\right\}$ is the superclick label assignment ($l_{s_i}$ is the binary superclick label for superpixel $s_i$), $\mathcal{N}(S)$ is the set of pairs of neighbor superpixels (that is, having part of boundary in common; we will also use the notation $\mathcal{N}(s)$ to denote the set of neighbors of the single superpixel $s$), and $\alpha_1$ is a weighing factor.

Unary potential $V_1$ models whether superpixel $s$ is likely to be a superclick or not. This ``likeliness'' depends on the number and quality of clicks inside the superpixel's region and on the vicinity to clicked superpixels\footnote{The reader might think that ``vicinity to clicked superpixels'' should be modeled as a pairwise potential, rather than unary. In fact, it should be modeled as unary because it is not an indication of whether two elements should be assigned the same label (which is what pairwise potentials represent); instead, it uses local information to indicate whether that item, \emph{individually}, is more likely to be assigned to a specific label ($1$ for ``superclick'' or $0$ for ``not superclick'')}. Therefore, $V_1$ is given by two contributions:
\begin{itemize}
 \item \textbf{\emph{Clickedness} $K_s$}: the more (high-quality) clicks a superpixel has received, the more it is likely to be a good candidate superclick. The clickedness score $K_s$ for superpixel $s$ is:
 \begin{equation}
  K_s = \underbrace{\frac{\left|C \cap s\right|}{\max\limits_{t \in S} \left| C \cap t\right|}}_\text{(\ref{eq:superclick_energy_clickedness}a)} \underbrace{\frac{1}{\left|C \cap s\right|} \sum_{c \in C \cap s} q_c}_\text{(\ref{eq:superclick_energy_clickedness}b)} = \frac{\sum\limits_{c \in C \cap s} q_c}{\max\limits_{t \in S} \left|C \cap t\right|}
  \label{eq:superclick_energy_clickedness}
 \end{equation}
 where $C \cap s$ is the set of clicks hitting superpixel $s$ and $\left|\cdot\right|$ is set cardinality. The first (unreduced) version explains more clearly what this formula is meant for: Eq. term (\ref{eq:superclick_energy_clickedness}a) indicates how many clicks, superpixel $s$ contains with respect to the superpixel containing most clicks in the processed frame; Eq. term (\ref{eq:superclick_energy_clickedness}b) is, instead, the average quality of clicks inside $s$, and encodes quality information in the score. The way this score is computed thus addresses items 1 and 2 of the above design principles.
 \item \textbf{Proximity to clicked superpixels $V_s$}: if $s$ has not received many clicks but is close to superpixels which did, we might want to take it into consideration as a potential superclick.  
 Of course, being close to clicked superpixels by itself is not enough: any superpixel just outside an object's boundary satisfies this requirement; this issue will be addressed by the pairwise potential $V_2$.\\
 Our proximity score $V_s$ is computed as the fraction of neighbor superpixels with clickedness score $K_{s_n} > 0.5$, with $s_n \in \mathcal{N}(s)$:
 \begin{equation}
  V_s = \frac{\left| \left\{ s_n \in \mathcal{N}(s) : K_{s_n} > 0.5 \right\} \right|}{\left|\mathcal{N}(s)\right|}
 \end{equation}
 
 Analogously, if $s$ gets enough clicks but is isolated, $V_s$ will be low and  $K_s$ won't suffice to label it as a superclick. Thus, $V_s$ balances items 3 and 4 of our design principles.
 
 \item \textbf{Unclicked regularizer}: the point of introducing the $V_s$ score is to allow a superpixel with few or no clicks to be labeled as superclick if its neighborhood hints that it should; however, if an unclicked superpixel is not adjacent to any clicked superpixels, its $V_1$ potential is zero, which is something we want to avoid. Consider, for example, the case of an object consisting of a large uniform region with a non-uniform users' click distribution (which is actually often the case, as users tend to click at the center of objects): by setting unclicked superpixels to a low (but not null) potential, we allow labels to ``spread'' from superclicks (as per item 3 of our design principles above)---as long as uniformity requirements, defined by potential $V_2$, apply.\\
 For this reason, we add a constant $U_s$ term to the $V_1$ potential, which should be small enough not to ``push'' too much toward the ``superclick'' label (since clickedness and vicinity clues suggest it should not be), but not so small that it cannot ever be labeled as such.\\
\end{itemize}

The definitions of $K_s$, $V_s$ and $U_s$ have been chosen so that the sum of those terms (clipped to 1 if necessary) can be interpreted as the probability that superpixel $s$ belongs to class ``superclick'', $P_{s,1} = P(l_s = 1 | C,S) = \min\left(K_s + V_s + U_s, 1\right)$. Similarly, the complementary probability $P_{s,0} = P(l_s = 0 | C,S) = 1 - P_{s,1}$ is the probability that $s$ is ``not a superclick''. In the energy function, $V_1$ is meant to represent the cost of assigning a certain label to each superpixel: such costs can be computed as the negative log-likelihood of the two probabilities above:
\begin{equation}
 V_1(s, l_s, C) =
 \begin{cases}
  -\log P_{s,1} & \text{if}~l_s = 1 \\
  -\log P_{s,0} & \text{if}~l_s = 0
 \end{cases}
\end{equation}

Pairwise potential $V_2$ is the cost of assigning different labels to two adjacent superpixels $s_1$ and $s_2$: ideally, it should be large for ``similar'' superpixels (so that they are assigned the same label) and small for superpixels for which no evidence exists that they should belong to the same class. Although in general this function could depend on the specific labels being assigned (so that, for example, the cost of assigning labels $(l_{s_1} = 1, l_{s_2} = 0)$ might be different than the cost of assigning labels $(l_{s_1} = 0, l_{s_2} = 1)$), in our case we focus only on estimating the optimal separation point between the ``superclick''/``non-superclick'' regions, based on visual similarity.\\
Therefore, potential $V_2$ is simply expressed as follows:
\begin{equation}
 V_2(s_1,s_2, l_{s_1}, l_{s_2}) = \text{exp}\left[-\beta_1 \chi^2(H_{s_1}, H_{s_2})\right]\mathcal{I}(l_{s_1} \neq l_{s_2})
\end{equation}
where $\chi^2(\cdot,\cdot)$ is the Chi-square distance, $H_{s_i}$ is the RGB color histogram of superpixel $s_i$, $\beta_1$ is a constant, and $\mathcal{I}$ is an indicator function which returns 1 if the arguments is true, and 0 otherwise (this ensures that $V_2$ is a submodular function, thus making the whole energy function graph-representable~\cite{Kolmogorov2004what}). Using a simple similarity measure such as the color histogram has a twofold justification: 1) by construction, superpixels have very little internal structure, so using more complex descriptors is unnecessary; 2) since the function has to be evaluated for all pairs of adjacent superpixels, it is important to perform as efficient operations as possible, in order to keep computation times reasonable.

\begin{figure}
\centering
\includegraphics[width=.24\textwidth]{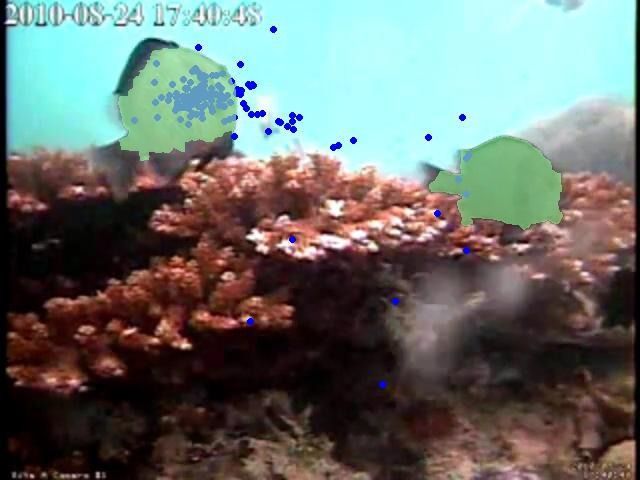}
\includegraphics[width=.24\textwidth]{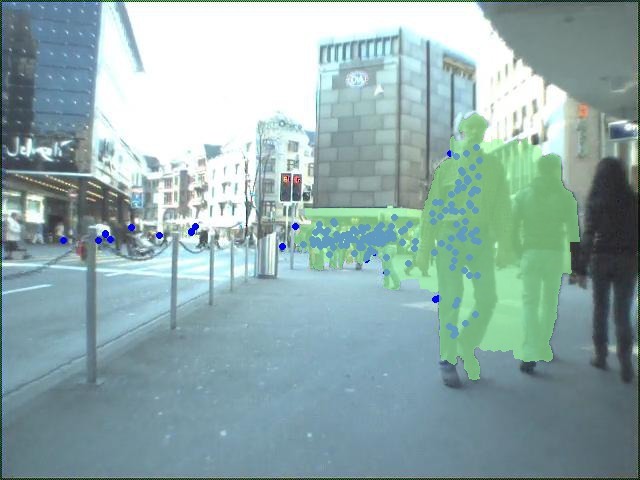} \\ \vspace{0.2cm}
\includegraphics[width=.24\textwidth]{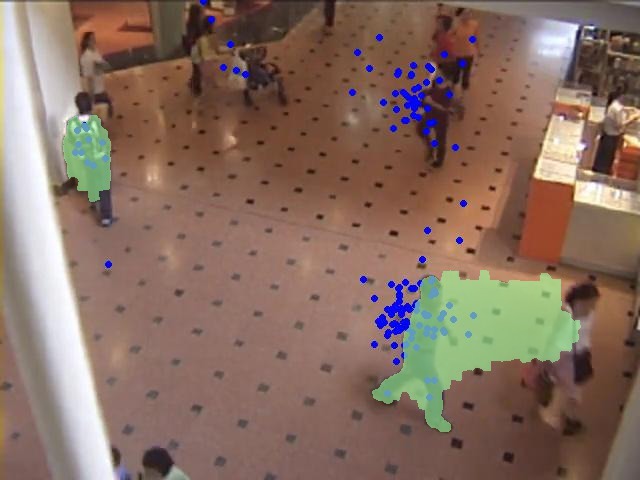}
\includegraphics[width=.24\textwidth]{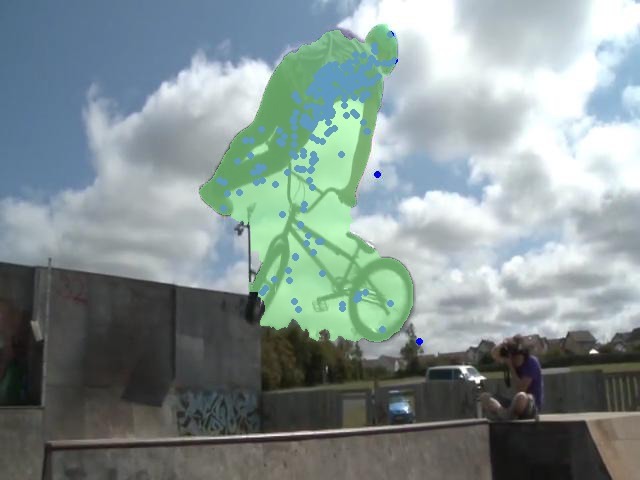}\\ \vspace{0.2cm}
\includegraphics[width=.24\textwidth]{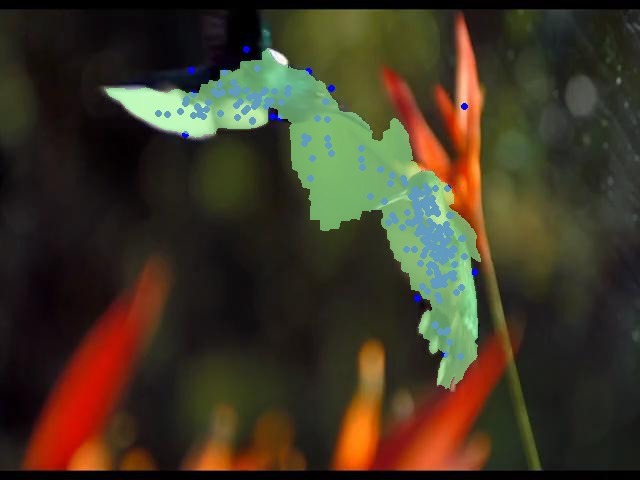}
\includegraphics[width=.24\textwidth]{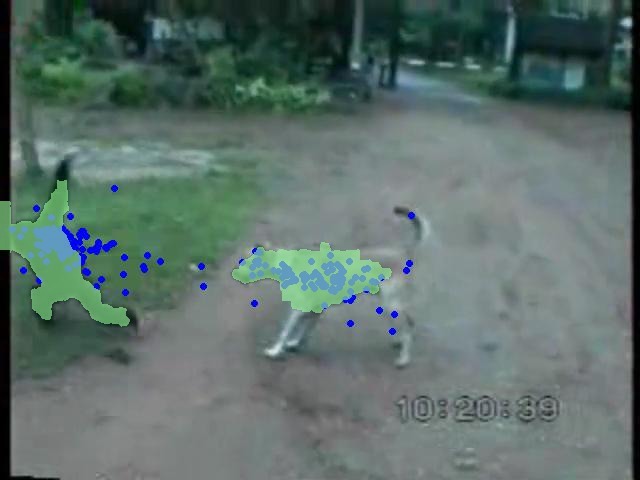}
\caption{Output examples for superclick identification: blue dots are users' clicks while green regions show the yielded segmentation masks. Segmentation refinement is carried out by including temporal constraints.}
\label{fig:superclicks}
\end{figure}

Once $E_1(\mathcal{L})$ has been minimized by means of graph cut, the extracted superclicks already provide a good approximated segmentation of the objects of interest in the scene, as shown by the examples in Fig.~\ref{fig:superclicks}. Nevertheless, output images at this stage can show segmentation errors, e.g., holes, oversegmentations, etc, and further processing by taking into account motion information is carried out to refine the obtained segmentation masks.

\subsection{Temporal Smoothing}
\label{sec:temporal}

The superclick extraction step turns a set of noisy clicks into a set of spatial coherent superclicks per frame, but ignores any temporal information which, instead, is necessary in video streams. 
Therefore, the next step for segmentation refinement consists in exploiting the \emph{temporal consistency} between consecutive frames to ``transfer'' labels across segmentations. 
The idea is that if a set of consecutive (in time) segmentations all mark a certain object as ``interesting'', then it is likely that they are correct; similarly, if no (or only few) segmentations include that object, it is probably safer to ignore it in the final output, especially if it is relatively isolated from other potential foreground objects. 
Two issues arise when trying to implement the above criterion: first, superclick segmentations are defined in terms of superpixel labels, and superpixel segmentation is not consistent in presence of motion; second, objects in a video typically move, thus the notion of ``a certain object'' across several frames implies the employment of a object/point tracking method. 

Our approach addresses both issues: 1) we define a \emph{temporal linking} between superclicks by extending the energy function, employed for superclick extraction, thus taking into account visual similarity between spatio-temporal regions; 2) we employ optical flow \cite{Liu2009opticalflow} to estimate where superpixels in frame  $t$ may have moved in frame $t+1$: in practice, we introduce pairwise potentials on all pairs of superpixels $\left\{s_t, s_{t+1}\right\}$ such that $s_t$ contains at least one pixel $p_t$ whose projection $p^{v_{p_t}}_{t\rightarrow {t+1}}=p_t+v_{p_t}$ into frame $t+1$ under the motion vector $v_{p_t}$ (i.e., $v_{p_t}$ is the motion vector computed between frame $t$ and frame $t+1$ for location $p_t$) is part of superpixel $s_{t+1}$ in frame $t+1$. Of course, it is unlikely that each superpixel will appear in only one such link, which allows to better ``explore'' the space around the estimated motion area, thus reducing the amount of error due to the optical flow and performing a more comprehensive analysis on the surrounding superpixels.\\
In the definition of the cost function employed for the temporal smoothing across frames $t-T$ and $t+T$, where $t$ is the current processed frame and $T$ is a constant which affects the number of frames involved in the temporal smoothing (i.e., $2T+1$), we assume to have identified superclicks for all the involved frames. In particular, we will refer to the same quantities as defined in Section~\ref{sec:superclick} and add an apex relative to the frame they refer to: for example, $l_s^t$ is the superclick label for superpixel $s$ in frame $t$, $S^{t+1}$ is the set of superpixels in frame $t+1$, and so on. The output label set will be identified by $\mathcal{L}$, and each label by $l_s$, without the temporal apex and they refer to the segmentation of the current processed frame. We can now introduce the energy function used for the final segmentation:
\begin{equation}
\begin{aligned}
 E_2(\mathcal{L}) = &  \sum_{\tau = t-T}^{t+T} \left[ \alpha_2 \sum_{s \in S^\tau} W_1(s, l_s, l_s^\tau) \right] +\\
 + & \sum_{\tau = t-T}^{t+T} \left[\sum_{(s_1,s_2) \in \mathcal{N}(S^\tau)} V_2(s_1,s_2, l_{s_1}, l_{s_2}) \right] + \\
+ & \sum_{(s_1,s_2) \in \mathcal{N}_T(\cup_{\tau = t-T}^{t+T} S^\tau)} V_2(s_1,s_2, l_{s_1}, l_{s_2})
\end{aligned}
\label{eq:time_energy}
\end{equation}


The first two lines of the cost function includes single-frame potentials, which consist of, respectively, unary potentials for each identified superpixel (first line) and pairwise potentials (second line) for each pair of superpixels belonging to the same frame. The last term (third line) enforces temporal smoothing, and consists of pairwise potentials computed over the set $\mathcal{N}_T(\cup_{\tau = t-T}^{t+T} S^\tau)$, which represents all pairs of superpixels (from all the frames in the considered time interval) satisfying the ``temporal linking'' described above, i.e., such that the two superpixels in each pair belong to temporally consecutive frames, and that at least one pixel belonging to one of them is projected onto the other by means of optical flow.

Similarly to $V_1$ (defined in Sect.~\ref{sec:superclick} ), unary potential $W_1$ models whether superpixel $s$ is more likely to be assigned to background or foreground \emph{per se}. In this stage, we simply assign a constant value to the potential depending on whether it had been identified, at the previous stage (see Sect.~\ref{sec:superclick}), as a superclick or not (i.e., depending on $l_s^\tau$).  In detail, given superpixel $s$, we set corresponding \emph{foreground cost} and \emph{background cost}; the value of each cost depends on $l_s^\tau$: if $s$ was labeled as a superclick, we expect it to be more likely that it is foreground, so the background cost should be higher, and vice versa. $W_1$ is therefore computed as follows:
\begin{equation}
 W_1(s, l_s, l_s^\tau) =
 \begin{cases}
  \gamma_1 & \text{if}~(l_s = 1 \land l_s^\tau = 1) \lor (l_s = 0 \land l_s^\tau = 0)\\
  \gamma_2 & \text{otherwise}
 \end{cases}
\end{equation}
with $\gamma_1 < \gamma_2$.

Pairwise potential $V_2$ is defined as in Section~\ref{sec:superclick}, but in the last term (third line of Formula~\ref{eq:time_energy}) of $E_2$ we employ it to evaluate the similarity not only between adjacent superpixels in the same frame, but also ``temporally-adjacent'' (according to $\mathcal{N}_T$) superpixels in consecutive frames. In order to deal with errors in optical flow computation, we do not simply assign a constant based on the presence or absence of a temporal link between two superpixels in consecutive frames, but also verify that they are visually similar and in fact refer to the same object/region in both frames. Thus, we manage to enforce the criteria according to which superpixels overlaying the same region across different frames should all be assigned the same label.

\begin{figure}
	\centering
	\includegraphics[width=.24\textwidth]{figures/mrf1/config_1_v_09_f_025_masked.jpg}
	\includegraphics[width=.24\textwidth]{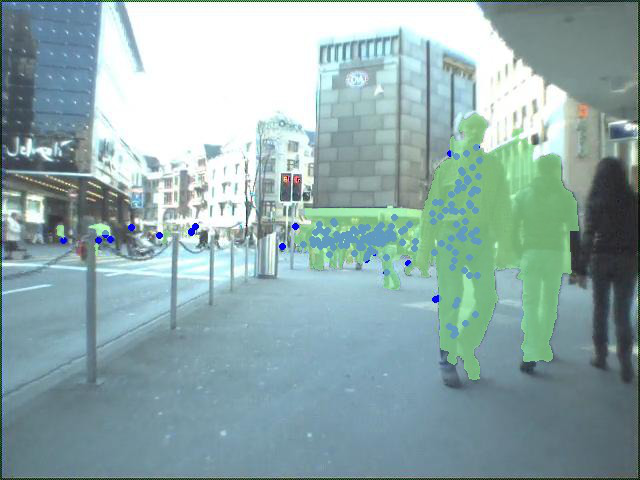}\\ \vspace{0.2cm}
	\includegraphics[width=.24\textwidth]{figures/mrf1/config_1_v_13_f_135_masked.jpg}
	\includegraphics[width=.24\textwidth]{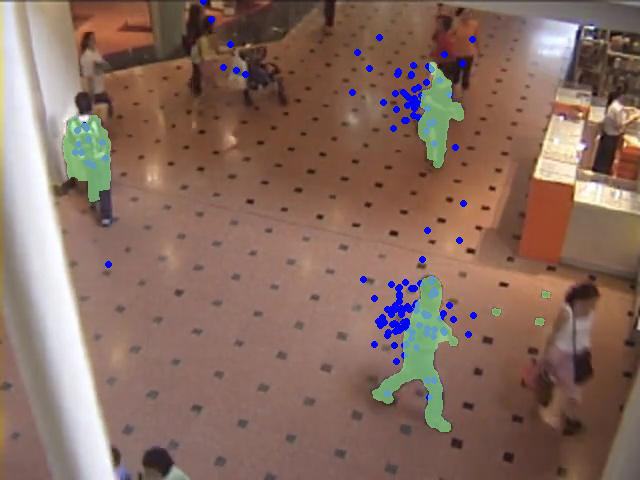}\\ \vspace{0.2cm}
	\includegraphics[width=.24\textwidth]{figures/mrf1/config_1_v_15_f_042_masked.jpg}
	\includegraphics[width=.24\textwidth]{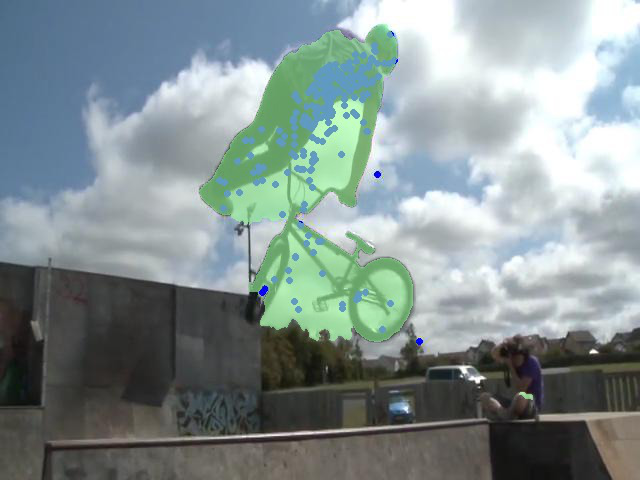}\\ \vspace{0.2cm}
	\includegraphics[width=.24\textwidth]{figures/mrf1/config_1_v_15_f_181_notime.jpg}
	\includegraphics[width=.24\textwidth]{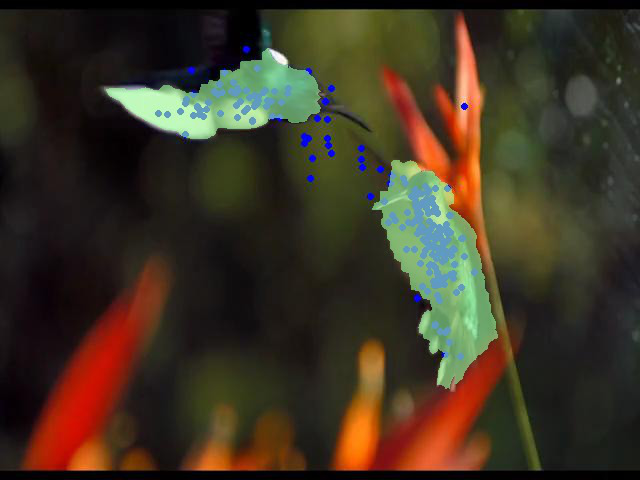}\\ \vspace{0.2cm}
	\includegraphics[width=.24\textwidth]{figures/mrf1/config_1_v_15_f_245_masked.jpg}
\includegraphics[width=.24\textwidth]{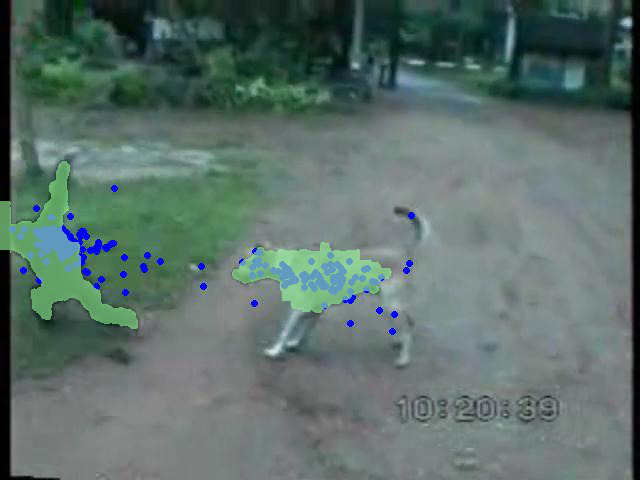}
	\caption{Qualitative comparison between segmentations obtained when excluding (first column, see Fig.~\ref{fig:superclicks}) and including temporal smoothing (second column).}
	\label{fig:superclicks1}
\end{figure}

Both $E_1$ and $E_2$ are binary pairwise energy functions with submodular pairwise potentials, and as such we minimize exactly them by graph-cuts in order to get the final segmentation for frame $t$.
Some qualitative examples are shown in Fig.~\ref{fig:superclicks1} (compared to those obtained by using only superclick extraction shown in Fig.~\ref{fig:superclicks}): it is easy to notice the difference in terms of segmentation quality achieved by analyzing a single frame only and by employing temporal smoothing, which is able to extract much better objects' shapes. It should also be noted that most of the processing (e.g. superpixel extraction and optical flow) can be shared when processing frames one after another, thus reducing the main processing time to superpixel segmentation and computation of optical flow for a single frame only.

\section{Experimental results}
\label{sec:results}
In this section we present the experimental results obtained by testing our gamification approach and link them to the state of the art on interactive video annotation and automated video object segmentation methods.

\subsection{Datasets}

For testing the accuracy of our method we created 7 game levels (each 300 frames long) from the following four datasets:
\begin{itemize}
 \item Fish dataset \cite{Spampinato2014texton}: we created two levels from the \emph{ComplexBkg1} and \emph{Standard1} video sequences, by choosing frame intervals with high density of ground truth frames and with several visible objects.
 \item I2R \cite{Li2003I2R}: we created two levels from the \emph{bootstrap} and \emph{shopping\_mall} videos, which are the most suitable ones for being used as game levels, since most videos in the dataset feature very few moving objects, in favor of dynamic background. The levels have been edited by selecting only parts with higher activity.
 \item ETH BIWI \cite{Ess2008ethbiwi}: we created two levels from the \emph{BAHNHOF} and \emph{JELMOLI} sequences created for urban multi-person tracking. These videos were particularly interesting because of the large number of moving targets appearing at the same time; the disadvantage was that annotations were provided only at the bounding-box level, which we nevertheless treated as pixel-wise segmentations (which causes the reported performance to be lower than they actually are, as our method performs pixel-level segmentation).
 \item SegTrack v2 \cite{Li2013segtrack2}: we created only one level by incorporating a subset of videos from the SegTrack v2 dataset (namely, \emph{birdfall}, \emph{bmx}, \emph{cheetah}, \emph{drift}, \emph{hummingbird}, \emph{monkey} and \emph{monkeydog}) into a single sequence, since many of them were just few dozen frames long. We favored sequences with multiple objects, and excluded videos where the target, though moving, appeared at the same frame location due to camera motion (e.g. the \emph{girl} sequence).
\end{itemize}

When we compared our methods to existing interactive video annotation methods and automated video object segmentation approaches we used the whole SegTrack v2.

\subsection{Collected data}

Our experiments involved only five players (two of them were children) outside of our research team, who were simply asked to compete with each other by achieving the highest possible score. The following information describes the amount of collected data and playing statistics:
\begin{itemize}
 \item \textbf{Level time: 30 seconds.}
 \item \textbf{Game time (7 levels): 3:30 minutes.}
 \item \textbf{75 games played}.
 \item \textbf{Total play time: 9:18 hours}. On average, each participant played for 1:52 hours, which may seem a lot. In fact, in a real public gaming scenario the amount of collected data corresponds to having about 160 users playing just one game, which is easily achievable by publishing the game with very little advertising on a social network.
 \item \textbf{Total number of clicks: 235,799}; the average number of clicks per frame was 52.4.
\end{itemize}

\subsection{Algorithm parameters}

In Sect.~\ref{sec:methods}, we introduced some parameters which control the trade-off between clicks and visual regularity in the segmentation process. We empirically set the values for those parameters, as follows: $\alpha_1 = 1/4$, $\alpha_2 = 1/5$, $U_s$ = 0.4, $\beta_1 = 5$, $T = 2 $, $\gamma_1 = 0.1$, $\gamma_2 = 0.9$ ($W_1$).

These parameter values were used to compute the results shown in the next section. It is important to note that the same parameters were used for all videos, although they have distinct differences in scenery, type of targets, motion patterns, motion speed, frame rate, etc. It is foreseeable that applying the same method to videos belonging to a more homogeneous set of videos would yield higher accuracy.

\subsection{Segmentation results}

This paper describes a interactive video object segmentation method based on spatio-temporal MRF optimization using the collaborative effort of multiple users. Thus, we first evaluated the role of temporal smoothing followed by the analysis of segmentation accuracy w.r.t to game play time, number of players, click delay and click quality. 
Then, we compared our method in terms of interaction times and segmentation accuracy with, respectively, recent state-of-the-art interactive video annotation methods and automated video object segmentation approaches.
The metrics employed for performance analysis were pixel-level precision ($Pr$), recall ($Rec$) and F-measure ($F_1$) and average Pascal Overlap Measure ($POM$: intersection over union between $T$ ground truth $GT$ masks  and output segmentations $S$) defined as:

\begin{equation}
\text{Pr} = \frac{\text{TP}}{\text{TP}+\text{FP}}
\end{equation}
\begin{equation}
\text{Rec} = \frac{\text{TP}}{\text{TP}+\text{FN}}
\end{equation}
\begin{equation}
\text{F}_1 = 2\cdot\frac{\text{Pr}\cdot \text{Rec}}{\text{Pr}+\text{Rec}}
\end{equation}

These performance metrics were computed by summing up the number of true positives, false positives and false negatives of each video category (i.e., without averaging across frames).

\begin{equation}
\text{POM(GT, S)} = \frac{1}{T}\sum_{i=1}^{T}\frac{|GT_i \cap S_i|}{|GT_i \cup S_i|}
\label{form:pom}
\end{equation}

\subsubsection{Role of spatio-temporal segmentation refinement}

Fig.~\ref{fig:superclicks1} shows a qualitative comparison in terms of segmentation outputs when employing  only superclick extraction phase (see Sect.~\ref{sec:superclick}) and when exploiting the temporal consistency between consecutive frames of superclicks. 
Table~\ref{res:temporal_smoothing} reports quantitatively how including spatio-temporal based refinement enhanced the segmentation accuracy. It can be noted that in some cases the accuracy gain was lower (ETH BIWI and I2R) than in others (Fish and SegTrack v2). This depends on the dynamics of the video sequences, i.e., in ETH BIWI and I2R the objects move slowly so users were able to click accurately on objects, while SegTrack v2 and Fish are characterized by  strong camera and object motion resulting in much noisier input data, and the spatio-temporal based refinement proved effective to recover users' failures in identifying objects.

The lower performance achieved in ETH BIWI and I2R than the ones in Fish and SegTrack v2, instead, can be explained by the number of objects in the scene. Indeed, some ETH BIWI and I2R videos depicted very crowded scenes and given the limited number of users playing the game, not all objects present in the scene were accurately identified.

For all the following evaluations, we used the method including the spatio-temporal segmentation refinement.

\begin{table*}
	\centering
	\caption{Average segmentation accuracy for the video categories employed in our game in terms of precision, recall and F-measure, when we employ only superclick extraction (first row) and when we refine the output segmentation by means of spatio-temporal linking between superclicks in consecutive frames (second row).}
	\begin{tabular}{lcccccccccccc}
		\toprule
		\multirow{2}{*}{\textbf{}} & \multicolumn{3}{c}{\textbf{Fish}} & \multicolumn{3}{c}{\textbf{ETH BIWI}} & \multicolumn{3}{c}{\textbf{I2R}}  & \multicolumn{3}{c}{\textbf{SegTrack v2}} \\
		\cmidrule(lr){2-4} \cmidrule(lr){5-7} \cmidrule(lr){8-10} \cmidrule(lr){11-13}
		{\bf Method} & \textbf{Pr} & \textbf{Rec} & \textbf{F${}_1$} & \textbf{Pr} & \textbf{Rec} & \textbf{F${}_1$} & \textbf{Pr} & \textbf{Rec} & \textbf{F${}_1$} & \textbf{Pr} & \textbf{Rec} & \textbf{F${}_1$} \\
		\cmidrule(lr){1-1} \cmidrule(lr){2-4} \cmidrule(lr){5-7} \cmidrule(lr){8-10} \cmidrule(lr){11-13}
			 Superclick extraction & 0.639	& 0.726 & \underline{0.680} & 0.597	& 0.629 & \underline{0.613} & 0.602 & 0.661 & \underline{0.6304} & 0.673 & 0.763 & \underline{0.716}\\
			Spatio-temporal refinement& 0.684 & 0.821 & \underline{0.746} & 0.647 & 0.636 & \underline{0.642} & 0.606 & 0.696 & \underline{0.6484} & 0.739 & 0.835 & \underline{0.784} \\
		\bottomrule
	\end{tabular}
	\label{res:temporal_smoothing}
\end{table*}

\subsubsection{Accuracy w.r.t. users' play time}
\label{sec:results_users_time}

Table~\ref{tab:overall_accuracy} shows how segmentation results vary in relation to the amount of users' play time in terms of pixel-level precision, recall and F-measure.

\begin{table*}
\centering
\caption{Average segmentation accuracy for the video categories employed in our game in terms of precision, recall and F-measure, for different values of cumulative users' play time.}
\begin{tabular}{lcccccccccccc}
\toprule
\multirow{2}{*}{\textbf{Playtime (hours)}} & \multicolumn{3}{c}{\textbf{Fish}} & \multicolumn{3}{c}{\textbf{ETH BIWI}} & \multicolumn{3}{c}{\textbf{I2R}}  & \multicolumn{3}{c}{\textbf{SegTrack v2}} \\
\cmidrule(lr){2-4} \cmidrule(lr){5-7} \cmidrule(lr){8-10} \cmidrule(lr){11-13}
& \textbf{Pr} & \textbf{Rec} & \textbf{F${}_1$} & \textbf{Pr} & \textbf{Rec} & \textbf{F${}_1$} & \textbf{Pr} & \textbf{Rec} & \textbf{F${}_1$} & \textbf{Pr} & \textbf{Rec} & \textbf{F${}_1$} \\
\cmidrule(lr){1-1} \cmidrule(lr){2-4} \cmidrule(lr){5-7} \cmidrule(lr){8-10} \cmidrule(lr){11-13}
9 h & 0.684 & 0.821 & \underline{0.746} & 0.647 & 0.636 & \underline{0.641} & 0.606 & 0.696 & \underline{0.648} & 0.739 & 0.835 & \underline{0.784} \\
8 h & 0.711 & 0.814 & \underline{0.759} & 0.632 & 0.670 & \underline{0.650} & 0.609 & 0.720 & \underline{0.660} & 0.721 & 0.827 & \underline{0.770} \\
7 h & 0.735 & 0.787 & \underline{0.760} & 0.672 & 0.612 & \underline{0.640} & 0.663 & 0.670 & \underline{0.666} & 0.770 & 0.819 & \underline{0.794} \\
6 h & 0.738 & 0.760 & \underline{0.749} & 0.655 & 0.644 & \underline{0.649} & 0.681 & 0.677 & \underline{0.679} & 0.758 & 0.796 & \underline{0.776} \\
5 h & 0.707 & 0.592 & \underline{0.644} & 0.689 & 0.564 & \underline{0.620} & 0.756 & 0.571 & \underline{0.650} & 0.816 & 0.758 & \underline{0.786} \\
3 h & 0.780 & 0.372 & \underline{0.504} & 0.716 & 0.472 & \underline{0.569} & 0.812 & 0.368 & \underline{0.507} & 0.883 & 0.617 & \underline{0.727} \\
1 h & 0.869 & 0.149 & \underline{0.254} & 0.816 & 0.339 & \underline{0.479} & 0.909 & 0.228 & \underline{0.365} & 0.929 & 0.450 & \underline{0.607} \\
\bottomrule
\end{tabular}
\label{tab:overall_accuracy}
\end{table*}

It is easy to notice that, at the moment we interrupted the experiment, the accuracy had already started to slightly decrease as the participants played more games (i.e., as more clicks were being collected). Indeed, it is intuitive that as soon as enough clicks have been collected which allow to sufficiently highlight superclicks from the background, additional clicks are more likely to be noisy than to actually cover any missing foreground region. Of course, in a large-scale application scenario with hundreds or thousands of videos, the amount of play time needed to reach this ``saturation point'' is much higher and would require a very large user base and/or users' time.

\subsubsection{Accuracy w.r.t. number of users}

Among the advantages of exploiting gamification to solve a complex or large-scale task, one of the most important is the variety of data patterns contributed by different users of the system. This is especially important in multi-target tasks such as segmentation with multiple objects, as single users tend to be biased to select the same object across multiple games. We evaluated this tendency by comparing the segmentation accuracy obtained by taking into consideration the clicks from a single user (single-player scenario) with the accuracy obtained by the other four players (many-players scenario). The chosen user for the single-player scenario is the one who played most games (23); in order to compute the accuracy for the many-players scenario, we randomly sampled several sets of clicks such that each set had the same number of clicks per video as the chosen single player and averaged the results. Table~\ref{tab:num_players_accuracy} show the comparison between the two scenarios.

\begin{table*}
\centering
\caption{Average segmentation accuracy for the video categories employed in our game in terms of precision, recall and F-measure, for the single-player and the many-players scenarios.}
\begin{tabular}{lcccccccccccc}
\toprule
\multirow{2}{*}{\textbf{Number of players}} & \multicolumn{3}{c}{\textbf{Fish}} & \multicolumn{3}{c}{\textbf{ETH BIWI}} & \multicolumn{3}{c}{\textbf{I2R}}  & \multicolumn{3}{c}{\textbf{SegTrack v2}} \\
\cmidrule(lr){2-4} \cmidrule(lr){5-7} \cmidrule(lr){8-10} \cmidrule(lr){11-13}
& \textbf{Pr} & \textbf{Rec} & \textbf{F${}_1$} & \textbf{Pr} & \textbf{Rec} & \textbf{F${}_1$} & \textbf{Pr} & \textbf{Rec} & \textbf{F${}_1$} & \textbf{Pr} & \textbf{Rec} & \textbf{F${}_1$} \\
\cmidrule(lr){1-1} \cmidrule(lr){2-4} \cmidrule(lr){5-7} \cmidrule(lr){8-10} \cmidrule(lr){11-13}
Single player & 0.577 & 0.203 & \underline{0.301} & 0.562 & 0.286 & \underline{0.379} & 0.781 & 0.305 & \underline{0.438} & 0.797 & 0.493 & \underline{0.609} \\
Multiple players  & 0.917 & 0.261 & \underline{0.407} & 0.797 & 0.415 & \underline{0.546} & 0.902 & 0.300 & \underline{0.451} & 0.870 & 0.543 & \underline{0.668} \\
\bottomrule
\end{tabular}
\label{tab:num_players_accuracy}
\end{table*}

The comparison clearly shows that using clicks coming from many users yielded markedly better performance than having a single user gather the same amount of data. Of course, the reported performance is lower than the best accuracy in Table~\ref{tab:overall_accuracy} because the click sets on which Table~\ref{tab:num_players_accuracy} was computed amount to less than 3 hours' cumulative play time.

To test the importance of integrating data from several participants in multi-target tasks, we compared the accuracy of the single-player and many-players with respect to the number of objects in a frame, across all videos. The results in Table~\ref{tab:num_players_objects} show that the difference in accuracy increases with the number of objects in a frame, hinting that, indeed, the variability introduced by a higher number of players positively affects a multi-target task with respect to fewer people working on it.

\begin{table*}
\centering
\caption{Average segmentation accuracy in terms of precision, recall and F-measure, for the single-player and the many-players scenarios, with respect to the number of objects in a frame.}
\begin{tabular}{lcccccccccccc}
\toprule
\multirow{2}{*}{\textbf{Number of players}} & \multicolumn{3}{c}{\textbf{1-2 objects}} & \multicolumn{3}{c}{\textbf{3-5 objects}} & \multicolumn{3}{c}{\textbf{6+ objects}} \\
\cmidrule(lr){2-4} \cmidrule(lr){5-7} \cmidrule(lr){8-10} \cmidrule(lr){11-13}
& \textbf{Pr} & \textbf{Rec} & \textbf{F${}_1$} & \textbf{Pr} & \textbf{Rec} & \textbf{F${}_1$} & \textbf{Pr} & \textbf{Rec} & \textbf{F${}_1$} \\
\cmidrule(lr){1-1} \cmidrule(lr){2-4} \cmidrule(lr){5-7} \cmidrule(lr){8-10} \cmidrule(lr){11-13}
Single player & 0.798 & 0.458 & \underline{0.582} & 0.564 & 0.219 & \underline{0.315} & 0.687 & 0.200 & \underline{0.310} \\
Multiple players  & 0.868 & 0.502 & \underline{0.636} & 0.760 & 0.285 & \underline{0.415} & 0.852 & 0.290 & \underline{0.432} \\
\bottomrule
\end{tabular}
\label{tab:num_players_objects}
\end{table*}

\subsubsection{Accuracy w.r.t. click delay}

While most algorithm parameters have a strictly mathematical meaning, click delay (i.e., the number of frames by which we shift user clicks to take human reaction delay into account) is particularly interesting to analyze, as it can be an important design choice  which should be made by taking into consideration several factors such as user base, target speed, frame rate, etc. We evaluated segmentation accuracy for different click delay values, and the results are shown in Table~\ref{tab:click_delay}. The specific value we used for our evaluations (2) is the one yielding the best average F-measure score, although the other values are quite close. It is interesting to see that the chosen value is not the optimal one for all videos -- for example, it is not for the ETH BIWI and I2R videos: indeed, those are the video categories with the lowest dynamics, requiring fewer sudden reactions and thus making it easier for users to follow the objects.

\begin{table*}
\centering
\caption{Average segmentation accuracy for the video categories employed in our game in terms of precision, recall and F-measure, for different click delay values.}
\begin{tabular}{lcccccccccccc|c}
\toprule
\multirow{2}{*}{\textbf{Click delay}} & \multicolumn{3}{c}{\textbf{Fish}} & \multicolumn{3}{c}{\textbf{ETH BIWI}} & \multicolumn{3}{c}{\textbf{I2R}}  & \multicolumn{3}{c}{\textbf{SegTrack v2}} \\
\cmidrule(lr){2-4} \cmidrule(lr){5-7} \cmidrule(lr){8-10} \cmidrule(lr){11-13}
& \textbf{Pr} & \textbf{Rec} & \textbf{F${}_1$} & \textbf{Pr} & \textbf{Rec} & \textbf{F${}_1$} & \textbf{Pr} & \textbf{Rec} & \textbf{F${}_1$} & \textbf{Pr} & \textbf{Rec} & \textbf{F${}_1$} & \textbf{Average F${}_1$} \\
\cmidrule(lr){1-1} \cmidrule(lr){2-4} \cmidrule(lr){5-7} \cmidrule(lr){8-10} \cmidrule(lr){11-13}
0 & 0.613 & 0.743 & \underline{0.672} & 0.611 & 0.692 & \underline{0.649} & 0.613 & 0.692 & \underline{0.650} & 0.664 & 0.789 & \underline{0.721} & \underline{0.673} \\
1 & 0.667 & 0.794 & \underline{0.725} & 0.613 & 0.678 & \underline{0.644} & 0.600 & 0.711 & \underline{0.650} & 0.688 & 0.815 & \underline{0.746} & \underline{0.692} \\
2 & 0.684 & 0.821 & \underline{0.746} & 0.647 & 0.636 & \underline{0.641} & 0.606 & 0.696 & \underline{0.648} & 0.739 & 0.835 & \underline{0.784} & \underline{0.705} \\
3 & 0.667 & 0.791 & \underline{0.724} & 0.613 & 0.675 & \underline{0.643} & 0.598 & 0.711 & \underline{0.650} & 0.688 & 0.815 & \underline{0.746} & \underline{0.691} \\
\bottomrule
\end{tabular}
\label{tab:click_delay}
\end{table*}

\subsubsection{Accuracy w.r.t. click quality}

In order to measure the effectiveness of our click quality assessment approach, we evaluated segmentation accuracy using subsets of clicks belonging to different quality ranges (namely, [0--0.6[, [0.6--0.8[, [0.8--1]), as shown in Table~\ref{tab:quality_accuracy}. Unlike our tests concerning the variation of accuracy with respect to the number of players, we did not normalize the size of the click sets, as the three ranges were chosen so that they would approximately contain the same number of clicks. It can be seen that our measurement for click quality reflects the accuracy of the resulting segmentations; it should be noted that, as in Table~\ref{tab:num_players_accuracy}, the reported results are sensibly lower than the best results (see Table~\ref{tab:overall_accuracy}) due to a lower number of used clicks. This demonstrates that click quality is necessary to achieve accurate segmentantions but the number of clicks is more important, i.e., it's more important to have many users than few high quality ones playing the game. 

\begin{table*}
\centering
\caption{Average segmentation accuracy for the video categories employed in our game in terms of precision, recall and F-measure, for different ranges of click quality. Along with each quality range, we report the fraction of clicks included in that range with respect to the total number of clicks.}
\begin{tabular}{lcccccccccccc}
\toprule
\multirow{2}{*}{\textbf{Quality range}} & \multicolumn{3}{c}{\textbf{Fish}} & \multicolumn{3}{c}{\textbf{ETH BIWI}} & \multicolumn{3}{c}{\textbf{I2R}}  & \multicolumn{3}{c}{\textbf{SegTrack v2}} \\
\cmidrule(lr){2-4} \cmidrule(lr){5-7} \cmidrule(lr){8-10} \cmidrule(lr){11-13}
& \textbf{Pr} & \textbf{Rec} & \textbf{F${}_1$} & \textbf{Pr} & \textbf{Rec} & \textbf{F${}_1$} & \textbf{Pr} & \textbf{Rec} & \textbf{F${}_1$} & \textbf{Pr} & \textbf{Rec} & \textbf{F${}_1$} \\
\cmidrule(lr){1-1} \cmidrule(lr){2-4} \cmidrule(lr){5-7} \cmidrule(lr){8-10} \cmidrule(lr){11-13}
0.8 -- 1   (32.7\%) & 0.817 & 0.423 & \underline{0.558} & 0.777 & 0.477 & \underline{0.591} & 0.872 & 0.391 & \underline{0.540} & 0.896 & 0.633 & \underline{0.742} \\
0.6 -- 0.8 (32.9\%) & 0.771 & 0.415 & \underline{0.540} & 0.762 & 0.482 & \underline{0.590} & 0.841 & 0.389 & \underline{0.532} & 0.872 & 0.640 & \underline{0.738} \\
0   -- 0.6 (34.4\%) & 0.392 & 0.223 & \underline{0.284} & 0.370 & 0.267 & \underline{0.310} & 0.407 & 0.319 & \underline{0.358} & 0.493 & 0.335 & \underline{0.399} \\
\bottomrule
\end{tabular}
\label{tab:quality_accuracy}
\end{table*}

\begin{table*}
	\centering
	\caption{Comparison in terms of segmentation accuracy - measured as average POM in percentage, respectively, achieved within the first 50 secs of annotation (first row) and maximun value (with the related interaction times)- between our approach and other interactive video annotation methods on a subset of 10 frames extracted from SegTrack v2 video sequences}
	\begin{tabular}{lcccc}
		\toprule
		& \cite{Sch13} & \cite{jain2014}   & \cite{NSB15} & Our method\\
		\midrule
		POM within 50 secs & 39.8 & 42.2 & 61.5 & {\bf 72.4} \\
		Maximum POM &56.6 ($\sim$1,200 secs) & 65.2 ($\sim$500 secs) & 84.3 ($\sim$1,400 secs)& {\bf 72.4} ($\sim$50 sec) \\
		\bottomrule
	\end{tabular}
	\label{tab:interactive}
\end{table*}

\subsubsection{Comparison with the state of the art interactive video annotation methods}

\begin{figure*}
	\centering
	\includegraphics[width=.68\textwidth]{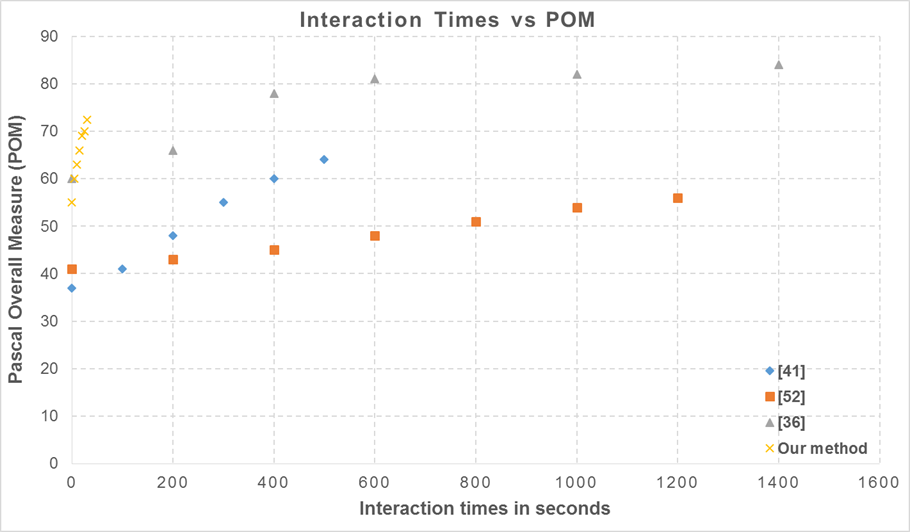}
	\caption{Interaction times vs segmentation accuracy. The figure shows that with the proposed approach we get a fairly good segmentation quality just after 30 seconds. When we allowed users to spend more time on the annotation task, the method in~\cite{NSB15} achieved the best performance with a POM of about 0.85 with an interaction time of about 1,400 seconds.}
	\label{fig:interaction_times}
\end{figure*}

We also compared our method to existing interactive video segmentation ones~\cite{Sch13,jain2014,NSB15} in terms of interaction times and accuracy. It has to be noted that for a single user the interaction time of our method depends only on the video length, i.e., there is a linear dependency between number of annotated frames and annotation times, while existing interactive methods~\cite{Sch13,jain2014,NSB15} show a non-linear (exponential-like) dependency. Nevertheless, using data generated by a single user in a single played game  would result in poor accuracy performance (as discussed in the previous section), thus we consider, for comparison with the state of the art, the cumulative time spent by a single user in multiple game sessions (generally one frame is shown for 0.2 seconds in our game, thus if we select the clicks of one user in five game sessions, the interaction time would be of 1 second.). 
We selected randomly 10 frames from SegTrack V2 and we assessed how the segmentation accuracy changed w.r.t. to the interaction times. For our method, we considered the clicks of the user who played most games in several game sessions. Fig.~\ref{fig:interaction_times} shows the interaction times of users against the achieved segmentation quality for our method and 
~\cite{Sch13,jain2014,NSB15}. Our approach yielded a reasonably accuracy over the 10 considered frames after approximately 50 seconds of playtime (i.e., about 20 game sessions) while the other approaches needed much more time. Quantitatively, Table~\ref{tab:interactive} reports the achieved segmentation accuracy (expressed by POM in percentage) over the 10 considered frames by all the comparing methods in two cases: a) within 50 seconds of annotation and b) the maximum achieved accuracy.
Let us recall that the interaction time (50 seconds corresponding to about 20 game sessions) for our method is given by the playtime of a single user, and that the same value can be achieved by involving 20 users, in parallel, playing only one game session, i.e., with a very little human annotation effort as opposed to the existing other solutions ~\cite{Sch13,jain2014,NSB15}.

\subsubsection{Comparison with the state of the art automated video segmentation methods}

Despite the proposed method is more inline with the research on interactive video annotation, it can be seen as a video object segmentation approach (with very little human intervention) and as such it is useful to link its performance with the state of the art on the automated methods. 
The comparison was again performed on the SegTrack v2 dataset (largely employed as a benchmarking dataset for video object segmentation), and we selected as comparing methods those ones posing the video object segmentation task as a superpixel labeling problem  using spatio-temporal MRF optimization, namely, \cite{Papazoglou2013fast,eccv,Giordano2015superpixel}. 
The results, in terms of average POM in percentage, are reported in Table~\ref{tab:segtrack}.

\begin{table}
	\centering
	\caption{Comparison in terms of segmentation accuracy - measured as average POM in percentage- between our approach and automated video object segmentation methods on SegTrack v2}
	\begin{tabular}{lcccc}
		\toprule
		& \cite{Papazoglou2013fast} & \cite{eccv} & \cite{Giordano2015superpixel} & Our method\\
		\midrule
		Average POM & 53.5 & 59.3 & 64.4  & {\bf 71.7} \\
		\bottomrule
	\end{tabular}
	\label{tab:segtrack}
\end{table}

\begin{figure*}
	\centering
	\includegraphics[width=.33\textwidth]{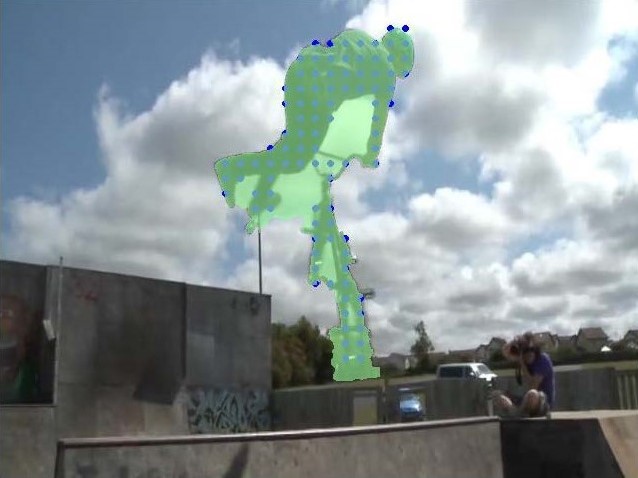}
	\includegraphics[width=.33\textwidth]{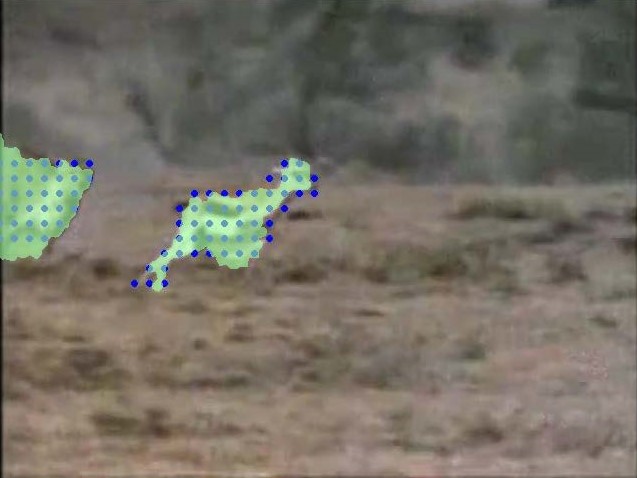}
	\includegraphics[height=4.5cm,width=.33\textwidth]{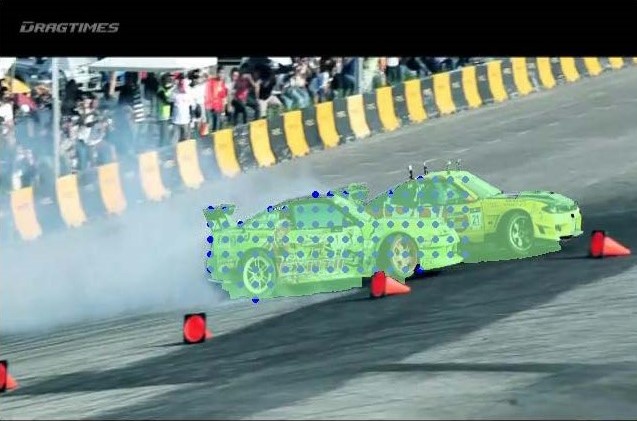} \\ \vspace{0.2cm}
	\includegraphics[height=4.5cm,width=.33\textwidth]{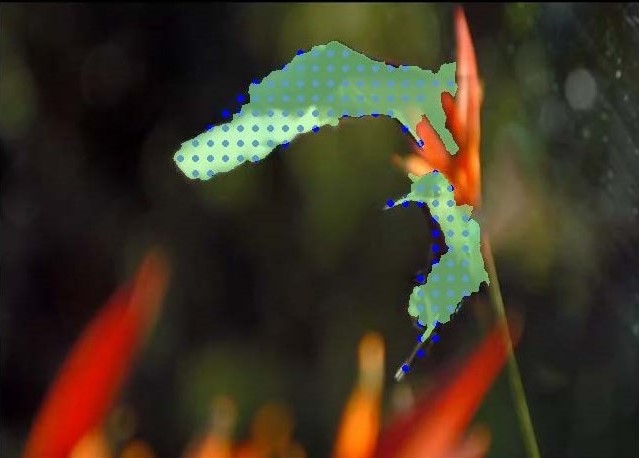}
	\includegraphics[width=.33\textwidth]{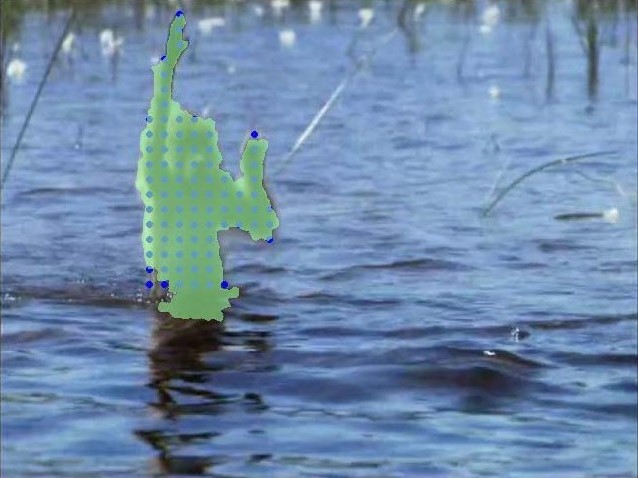}
	\includegraphics[width=.33\textwidth]{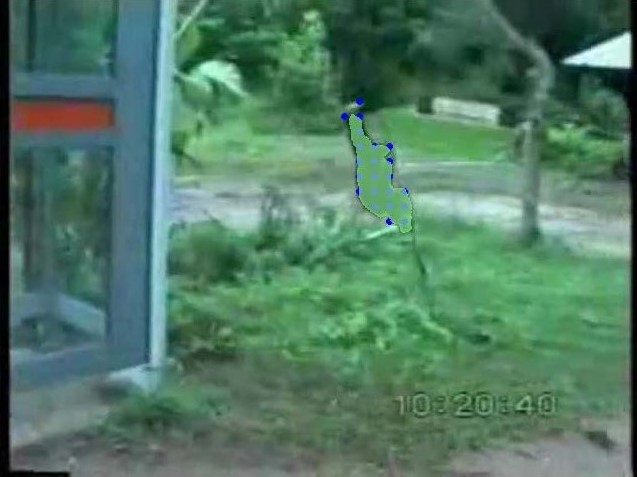}
	\caption{Output segmentations when using only points within objects (i.e., taken from ground truth segmentation masks) of interest: blue dots are ground truth points while green regions show the yielded segmentation masks.}
	\label{fig:good_gt}
\end{figure*}

Tables~\ref{tab:interactive} and \ref{tab:segtrack} indicate that our method performs better than automated video object segmentation methods and slightly worse than interactive video annotation approaches. This is not surprising since interactive video annotation tools require users to spend more time in providing accurate annotation but are barely usable in case of large video datasets. This makes our method a very good trade-off between accuracy and annotation times.
However, when we used only points (uniformly taken from ground truth masks) within objects of interest (this is might be a typical interactive video annotation scenario where users might be asked to accurately select foreground pixels) we obtain a much higher accuracy with an average POM of about 0.85 as shown in Fig.~\ref{fig:good_gt}. 

\begin{figure*}
	\centering
	\includegraphics[width=.4\textwidth]{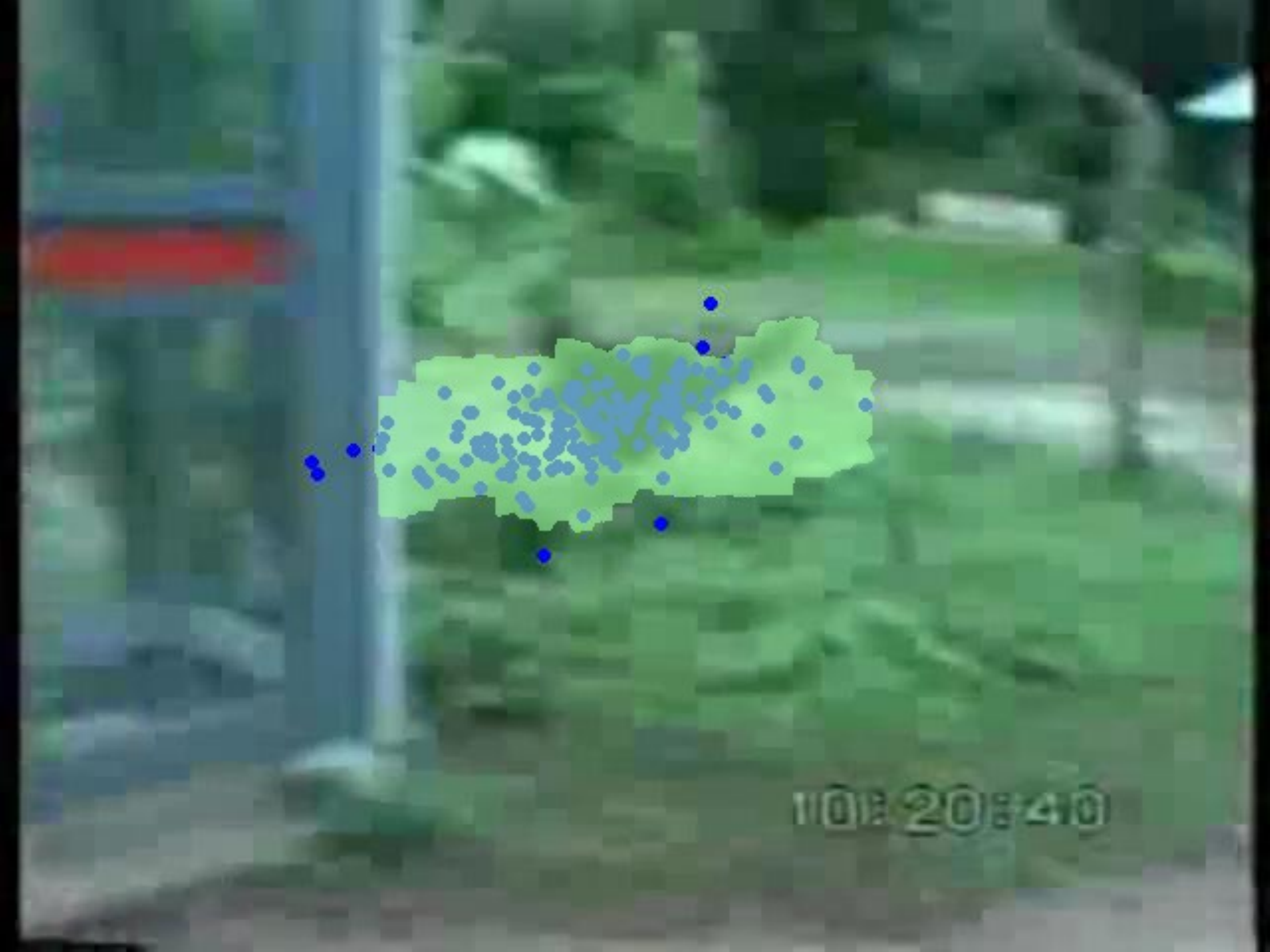}
	\includegraphics[width=.4\textwidth]{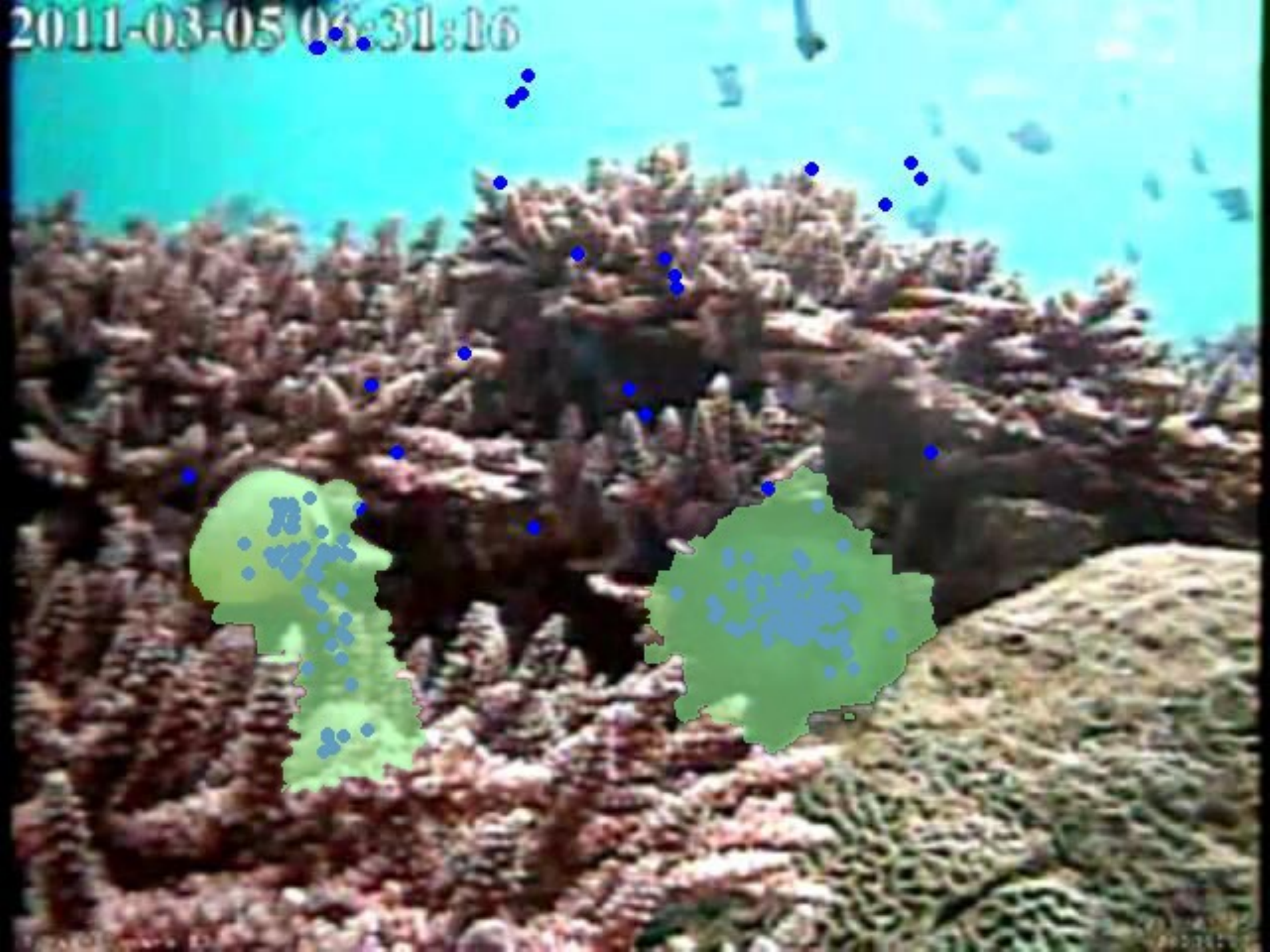}\\ \vspace{0.2cm}
	\includegraphics[width=.4\textwidth]{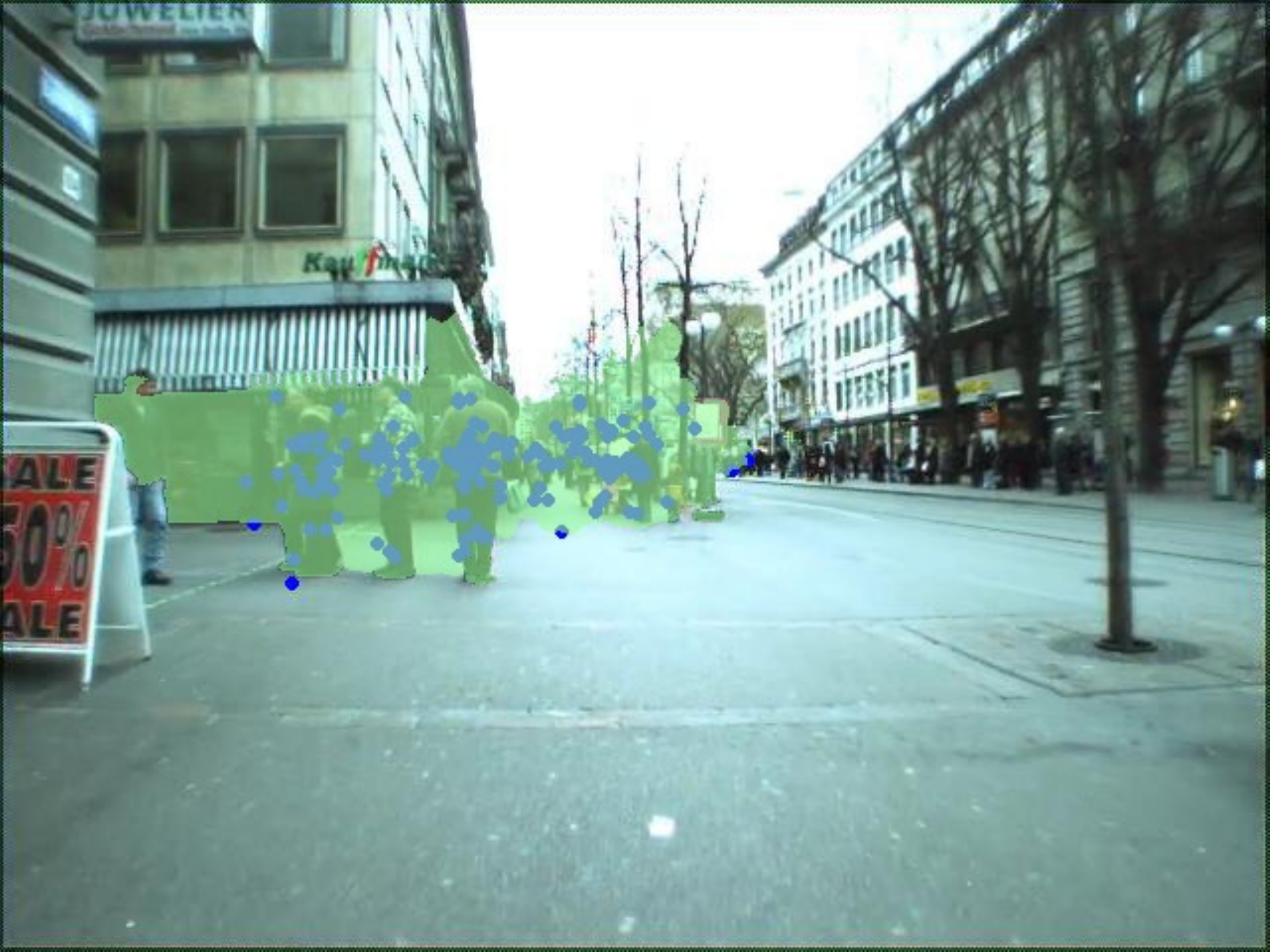}
	\includegraphics[width=.4\textwidth]{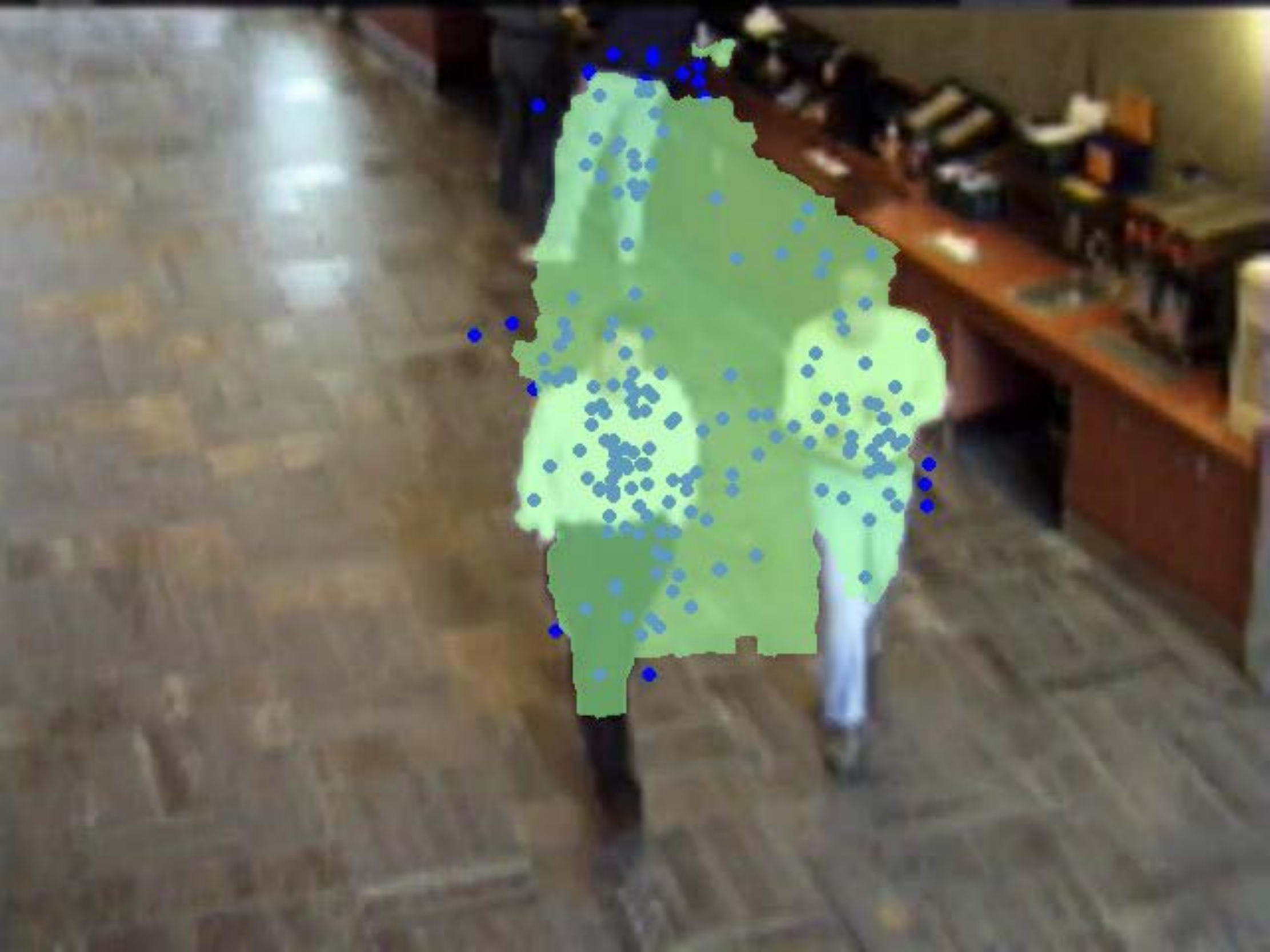}
	\caption{Failure cases: User clicks were extremely inaccurate resulting in wrong object segmentations.}
	\label{fig:sample_results}
\end{figure*}

Fig.~\ref{fig:sample_results} shows some failure cases of the proposed method: user clicks diverged substantially from objects' position hitting also background regions that were then classified as foreground. Our spatio-temporal refinement module was not able to recover such failures since in the previous phase (i.e., superclick extraction) several wrong superpixels were marked as superclicks. 

\subsubsection{Running times}

Using $T = 2$, processing a single frame requires solving five MRFs for superclick extraction and one temporal MRF for accurate segmentation. Our Matlab implementation, run on a PC with a quad-core i7 CPU and 8 GB RAM, takes 3 seconds for superclick extraction in a single frame and 30 seconds for the temporal MRF (actually, for the four optical flow computations which link superpixels in time; MRF solving time is negligible), which would amount to 45 seconds in total. However, after the initial bootstrap phase, segmenting a new frame can benefit from already-extracted superpixels and computed the optical flow for previous frames, so processing is reduced to a single superpixel extraction and a single optical flow computation, resulting in a frame processing time of 10.5 seconds.

\subsubsection{Available Resources}

In the authors' webpage, the source code for object segmentation taking user clicks as input together with the videos used as game levels are available. We do not release the source code for the game, instead, we provide a web service where interested people can set up their game using their own videos and get the output segmentations as well as raw data (e.g., user clicks).

\section{Concluding Remarks}
\label{sec:conclusions}
In this paper we have described an interactive video object segmentation method 
which combines effectively games with a purpose strategy with collaborative human  efforts. The performance analysis showed that our method outperforms in terms of segmentation accuracy state-of-the-art automated video object segmentation methods and is more suitable for large scale video annotation than classic interactive video annotation tools. 
We also demonstrated how including spatio-temporal regularization enhances greatly performance than using only spatial information and user-provided hard constraints.  We also release the source code for human-guided video object segmentation as well as a web service that enables interested people to set up their game (with their own videos) and download user generated data. 
As future work, we plan to carry out experiments at a larger scale involving more users (as in~\cite{6595949}) and videos. In addition as inhibition of return and click delay are two important aspects of the approach, we plan to make them adaptive according to, respectively, play time (thus to enforce users to identify as many objects as possible) and video motion characteristics.

\bibliographystyle{elsarticle-num}

\end{document}